\documentclass[twoside,twocolumn,9pt]{article}
\usepackage{extsizes}
\usepackage[super,sort&compress,comma]{natbib} 
\usepackage[version=3]{mhchem}
\usepackage[left=1.5cm, right=1.5cm, top=1.785cm, bottom=2.0cm]{geometry}
\usepackage{balance}
\usepackage{mathptmx}
\usepackage{sectsty}
\usepackage{graphicx} 
\usepackage{lastpage}
\usepackage[format=plain,justification=justified,singlelinecheck=false,font={stretch=1.125,small,sf},labelfont=bf,labelsep=space]{caption}
\usepackage{float}
\usepackage{fancyhdr}
\usepackage{fnpos}
\usepackage[english]{babel}
\addto{\captionsenglish}{%
  
}
\usepackage{array}
\usepackage{droidsans}
\usepackage{charter}
\usepackage[T1]{fontenc}
\usepackage[usenames,dvipsnames]{xcolor}
\usepackage{setspace}
\usepackage[compact]{titlesec}
\usepackage{hyperref}

\usepackage{epstopdf}

\definecolor{cream}{RGB}{222,217,201}
\begin{document}

\pagestyle{fancy}
\thispagestyle{plain}
\fancypagestyle{plain}{
\renewcommand{\headrulewidth}{0pt}
}

\makeFNbottom
\makeatletter
\renewcommand\LARGE{\@setfontsize\LARGE{15pt}{17}}
\renewcommand\Large{\@setfontsize\Large{12pt}{14}}
\renewcommand\large{\@setfontsize\large{10pt}{12}}
\renewcommand\footnotesize{\@setfontsize\footnotesize{7pt}{10}}
\renewcommand\scriptsize{\@setfontsize\scriptsize{7pt}{7}}
\makeatother

\renewcommand{\thefootnote}{\fnsymbol{footnote}}
\renewcommand\footnoterule{\vspace*{1pt}%
\color{cream}\hrule width 3.5in height 0.4pt \color{black} \vspace*{5pt}} 
\setcounter{secnumdepth}{5}

\makeatletter 
\renewcommand\@biblabel[1]{#1}            
\renewcommand\@makefntext[1]%
{\noindent\makebox[0pt][r]{\@thefnmark\,}#1}
\makeatother 
\renewcommand{\figurename}{\small{Fig.}~}
\sectionfont{\sffamily\Large}
\subsectionfont{\normalsize}
\subsubsectionfont{\bf}
\setstretch{1.125} 
\setlength{\skip\footins}{0.8cm}
\setlength{\footnotesep}{0.25cm}
\setlength{\jot}{10pt}
\titlespacing*{\section}{0pt}{4pt}{4pt}
\titlespacing*{\subsection}{0pt}{15pt}{1pt}

\fancyfoot{}
\fancyfoot[LO,RE]{\vspace{-7.1pt}\includegraphics[height=9pt]{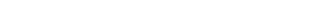}}
\fancyfoot[RO]{\footnotesize{\sffamily{1--\pageref{LastPage} ~\textbar  \hspace{2pt}\thepage}}}
\fancyfoot[LE]{\footnotesize{\sffamily{\thepage~\textbar\hspace{3.45cm} 1--\pageref{LastPage}}}}
\fancyhead{}
\renewcommand{\headrulewidth}{0pt} 
\renewcommand{\footrulewidth}{0pt}
\setlength{\arrayrulewidth}{1pt}
\setlength{\columnsep}{6.5mm}
\setlength\bibsep{1pt}

\makeatletter 
\newlength{\figrulesep} 
\setlength{\figrulesep}{0.5\textfloatsep} 

\newcommand{\topfigrule}{\vspace*{-1pt}%
\noindent{\color{cream}\rule[-\figrulesep]{\columnwidth}{1.5pt}} }

\newcommand{\botfigrule}{\vspace*{-2pt}%
\noindent{\color{cream}\rule[\figrulesep]{\columnwidth}{1.5pt}} }

\newcommand{\dblfigrule}{\vspace*{-1pt}%
\noindent{\color{cream}\rule[-\figrulesep]{\textwidth}{1.5pt}} }

\newcommand{\correction}[1]{\textcolor{blue}{#1}}

\makeatother

\twocolumn[
  \begin{@twocolumnfalse}
\begin{tabular}{m{0.cm} p{15.5cm} }

 & \noindent\LARGE{\textbf{LivePyxel: Accelerating image annotations with a Python-integrated webcam live streaming}} \\
 & \vspace{0.3cm} \\

 & \noindent\large{Uriel Garcilazo-Cruz,\textit{$^{a,b,\dag}$}, Joseph O. Okeme\textit{$^{a,b}$}, and Rodrigo A. Vargas-Hernández\textit{$^{a,b,c,\dag}$}} \\

\end{tabular}

 \end{@twocolumnfalse} \vspace{0.6cm}

  ]

\renewcommand*\rmdefault{bch}\normalfont\upshape
\rmfamily
\section*{}
\vspace{-1cm}


\author{
Uriel Garcilazo-Cruz$^{a,b\,\dag}$,
Joseph O. Okeme$^{a,b}$,
and Rodrigo A. Vargas-Hernández$^{a,b,c\,\dag}$
}


\sffamily{
    \textbf{The lack of flexible annotation tools has hindered the deployment of AI models in some scientific areas. Most existing image annotation software requires users to upload a precollected dataset, which limits support for on-demand pipelines and introduces unnecessary steps to acquire images. This constraint is particularly problematic in laboratory environments, where on-site data acquisition from instruments such as microscopes is increasingly common. In this work, we introduce \texttt{LivePixel}, a Python-based graphical user interface that integrates with imaging systems, such as webcams, microscopes, and others, to enable on-site image annotation. LivePyxel is designed to be easy to use through a simple interface that allows users to precisely delimit areas for annotation using tools commonly found in commercial graphics editing software. Of particular interest is the availability of Bézier splines and binary masks, and the software's capacity to work with non-destructive layers that enable high-performance editing. LivePyxel also integrates a wide compatibility across video devices, and it's optimized for object detection operations via the use of OpenCV in combination with high-performance libraries designed to handle matrix and linear algebra operations via Numpy effectively. LivePyxel facilitates seamless data collection and labeling, accelerating the development of AI models in experimental workflows. 
    LivePyxel is freely available at \url{https://github.com/UGarCil/LivePyxel}
    }
}


\rmfamily 

\footnotetext{\textit{$^{a}$~Department of Chemistry and Chemical Biology, McMaster University, Hamilton, ON, Canada\\
$^{b}$~School of Computational Science and Engineering, McMaster University, Hamilton, ON, Canada}}

\footnotetext {\textit{$^{c}$~Brockhouse Institute for Materials Research, McMaster University, Hamilton, ON, Canada\\
\dag~Corresponding author: garcilau@mcmaster.ca, vargashr@mcmaster.ca}}

\section{Introduction}
The capabilities of vision models (ViM) depend critically on the quality and availability of images\cite{BilalEtAl2024, chi2020deep,KARIMI2020101759,singh2024shifting,rs15204987,taran2020impact,djuravs2024dataset,taran2020impact}. However, collecting high-quality annotated images is time consuming and this annotation bottleneck \cite{litjens2017survey} negatively impacts the integration of ViM in highly specialized scientific domains, such as cellular imaging \cite{tajbakhsh2020embracing,moen2019deep}, civil infrastructure \cite{Yuan2024Review}, crop profile characterization \cite{agronomy15051157}, environmental microscopy \cite{Luo_et_al_2018} and marine conservation \cite{sauder2025coralscapes} to name a few. To address these data limitations, researchers have developed alternative approaches, including data augmentation schemes \cite{wang2021annotation} and more powerful models capable of segmenting and classifying diverse object types with minimal training (e.g., the Segment Anything Model (SAM)~\cite{kirillov2023segany}. However, these techniques still rely on the precision and accuracy of the initial annotations of domain experts. Ensuring high-quality specialized datasets requires the input of these experts \cite{Nodine1999}, which can be both a limiting factor and financially costly \cite{hosny2018artificial}.
The increasing availability of affordable, high-resolution imaging hardware highlights the importance of easy-to-use open-source software. Such tools can lower technical barriers and facilitate broader participation from individuals who possess domain knowledge but lack the expertise to collect and preprocess large datasets systematically.\\

In the fields of object detection and image segmentation, accessible annotation tools must be compatible with a wide range of imaging devices. We argue that many scientific workflows, particularly those involving microscopes or specimen curation, would benefit greatly from the ability to capture and annotate images in real time. For instance, navigating a microscope slide or processing large biological collections often involves domain experts who serve simultaneously as annotators. Separating the capture and annotation steps can interrupt this workflow and reduce efficiency.
In these contexts, the ability to annotate during image acquisition improves both efficiency and accuracy.

An image annotation tool focused on usability should also prioritize simple graphical user interface (GUI) components with a shallow learning curve, minimizing the need for technical support or costly training for specialized personnel. Although existing tools like LabelMe \cite{russell2008labelme}, VGG Image Annotator (VIA) \cite{dutta2019vgg}, and COCO Annotator \cite{cocoannotator} offer pixel-level annotations for segmentation tasks, they lack live camera integration, a critical gap for workflows requiring immediate feedback or iterative labeling. These tools, though flexible in formatting (e.g., COCO JSON, Pascal VOC), often require users to navigate feature-heavy interfaces or offline workflows (e.g., biologists, field technicians).
Moreover, most annotation software supports pixel-level annotation only through polygons or rectangular boxes. While effective for rigid geometries, these primitives are poorly suited for organic or curved structures. Approximating a smooth contour with polygons requires a high number of vertices, introducing annotation inefficiency and potential geometric bias that can propagate into downstream vision models. By contrast, graphic design software routinely uses Bézier splines to capture curves with minimal control points and high precision. This approach offers a more natural representation for biological or irregular shapes, making splines a compelling alternative to conventional polygon-based labeling.
Commercial platforms such as Labelbox and Supervisely emphasize collaboration features but omit both live annotation and spline support. RectLabel (macOS-only) supports Bézier curves but does not allow on-site input. Web-based tools like CVAT \cite{cvat_2023} and Label Studio \cite{labelstudio_2020} provide scalability yet remain restricted to pre-recorded media. Together, these limitations highlight the absence of an integrated solution that combines live annotation, spline-based precision, and lightweight usability.

We present LivePyxel, developed here, an open-source Python GUI that integrates on-site video device input with Bézier spline-based segmentation, enabling precise pixel-level annotation of curved structures. 
LivePyxel was initially developed for on-demand annotation of microscopy images, but it also supports any video device accessible via OpenCV and Python.
LivePyxel combines a lightweight, accessible interface with flexible annotation tools, including Bézier splines, polygons, and threshold-based masks, making it suitable for segmentation workflows across diverse research domains. The software is freely available at \url{https://github.com/UGarCil/LivePyxel} and can be installed through PyPI, with installation details found in the latest version of the repository along with tutorials, examples, and additional code. 

The paper is structured as follows: Section \ref{sec:livepyxel} describes the specification and deployment details of LivePyxel, developed in this work. Section \ref{sec:examples} demonstrates its application through an image segmentation task, where annotated images are used to train a ViM \cite{ronneberger2015u} in two scenarios: 1) segmenting eight different microorganisms and 2) performing data engineering with binary masks. LivePyxel is designed to streamline data collection and labeling, accelerating the development of AI models in experimental workflows that require on-site data manipulation or large-scale batch processing.

\section{LivePyxel}\label{sec:livepyxel}
LivePyxel is composed of an ecosystem of Python scripts that work cohesively to provide users with a graphical interface. This architecture provides a tool with a higher level of abstraction compared to most data science workflows, typically executed via Jupyter Notebooks.
The abstraction level is higher because the code that makes the software does not directly encode the annotation process, but rather builds interoperability across objects that make up the graphical user interface (GUI). This architecture was chosen because the domain of the task addressed by the program naturally belongs to a graphics editor, and because graphical interfaces make the tool available to a wider audience of users with very little experience in programming and code.

\begin{figure}[h]
  \centering
    \includegraphics[width=0.95\columnwidth]{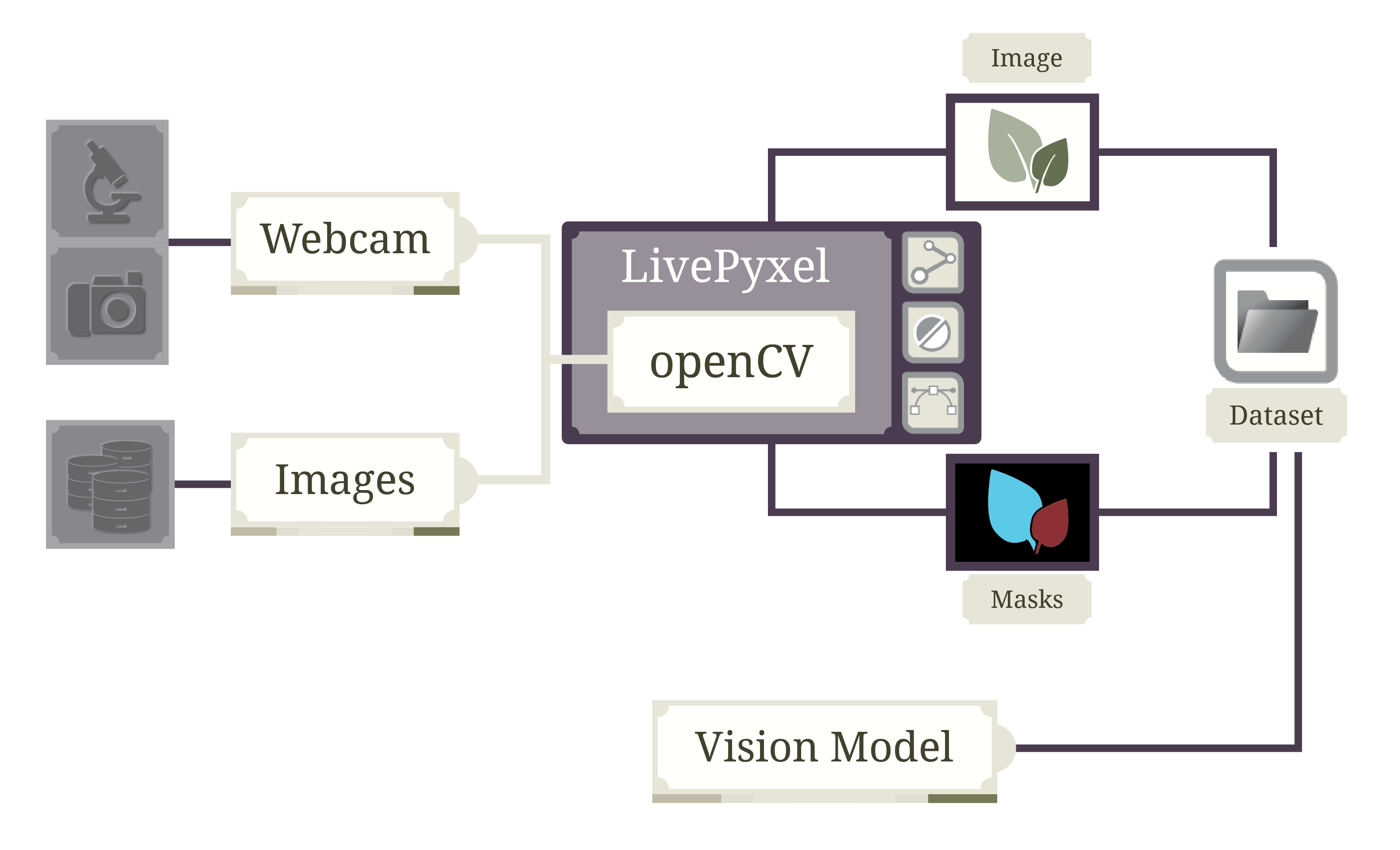}
    \caption{Overview of the LivePyxel architecture, illustrating the integration of webcam and image inputs with OpenCV for on-site image annotation. The system supports input from live microscopy or camera feeds as well as pre-existing datasets, processes them through the LivePyxel interface, and outputs both annotated images and segmentation masks for storage or further analysis by a vision model.}
    \label{fig:diagramFlow}
\end{figure}

The broad compatibility of LivePyxel with imaging devices is achieved through the use of the OpenCV \cite{bradski2000opencv} library, which enables on-site video input from virtually any camera and streams it directly to the annotation canvas. We chose OpenCV for its efficient handling of images as numerical matrices, leveraging a compiled back-end for performance while maintaining flexibility through seamless integration with NumPy \cite{harris2020array} for fast manipulation in Python. The GUI itself is built using the Qt framework, which allows NumPy arrays to be rendered directly as images, resulting in a responsive and user-friendly interface. Fig.~\ref{fig:diagramFlow} is an overview of the LivePyxel architecture.

\begin{figure*}[t]
  \centering
  \includegraphics[width=0.9\linewidth]{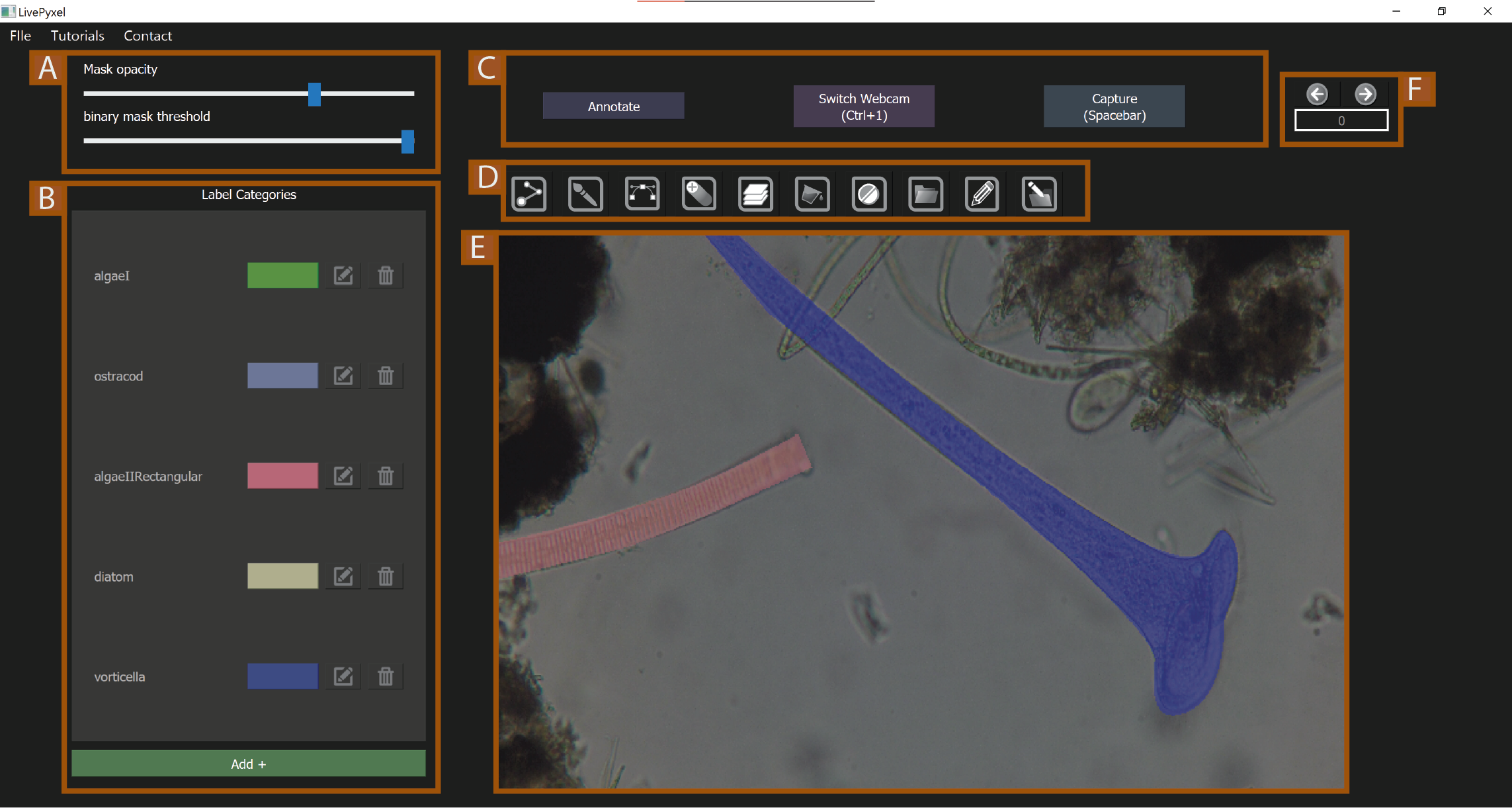}
  \caption{Preview of the LivePyxel graphical user interface (GUI) with its main components labeled: (A) sliders for adjusting mask opacity and binary mask threshold, (B) labels panel for managing annotation categories, (C) annotation panel with main control buttons, (D) toolbar for selecting drawing and editing tools, (E) central canvas where objects are annotated on microscopy images in real time, and (F) navigation controls for browsing through image frames or dataset entries.}
  \label{fig:livepyxel_gui}
\end{figure*}

The LivePyxel GUI integrates several interactive components to streamline the annotation workflow. Mask display properties, such as opacity and binary threshold, can be adjusted using the sliders in the control section (Fig.~\ref{fig:livepyxel_gui}-A). Annotation categories are managed in the labels panel (Fig.~\ref{fig:livepyxel_gui}-B), where users can add, edit, or delete classes. The annotation panel (Fig.~\ref{fig:livepyxel_gui}-C) provides high-level controls for switching input sources, capturing frames, and toggling annotation mode. Drawing and editing actions are performed using the toolbar (Fig.~\ref{fig:livepyxel_gui}-D), which offers tools for creating, modifying, or erasing masks. The canvas (Fig.~\ref{fig:livepyxel_gui}-E) displays either live microscope or uploaded images, over which masks are layered non-destructively in real time. Navigation buttons (Fig.~\ref{fig:livepyxel_gui}-F) allow users to cycle through frames or dataset entries. Once annotations are complete, LivePyxel saves both the annotated images and corresponding masks into automatically generated subfolders within a user-specified directory. In addition to on-site webcam annotation, LivePyxel also supports traditional workflows by allowing users to upload pre-existing image datasets that follow the required folder structure.

\subsection{Bézier splines for non-tradictional curvatures}\label{sec:Bézier_splines}
\begin{figure}[h]
  \centering
    \includegraphics[width=0.95\columnwidth]{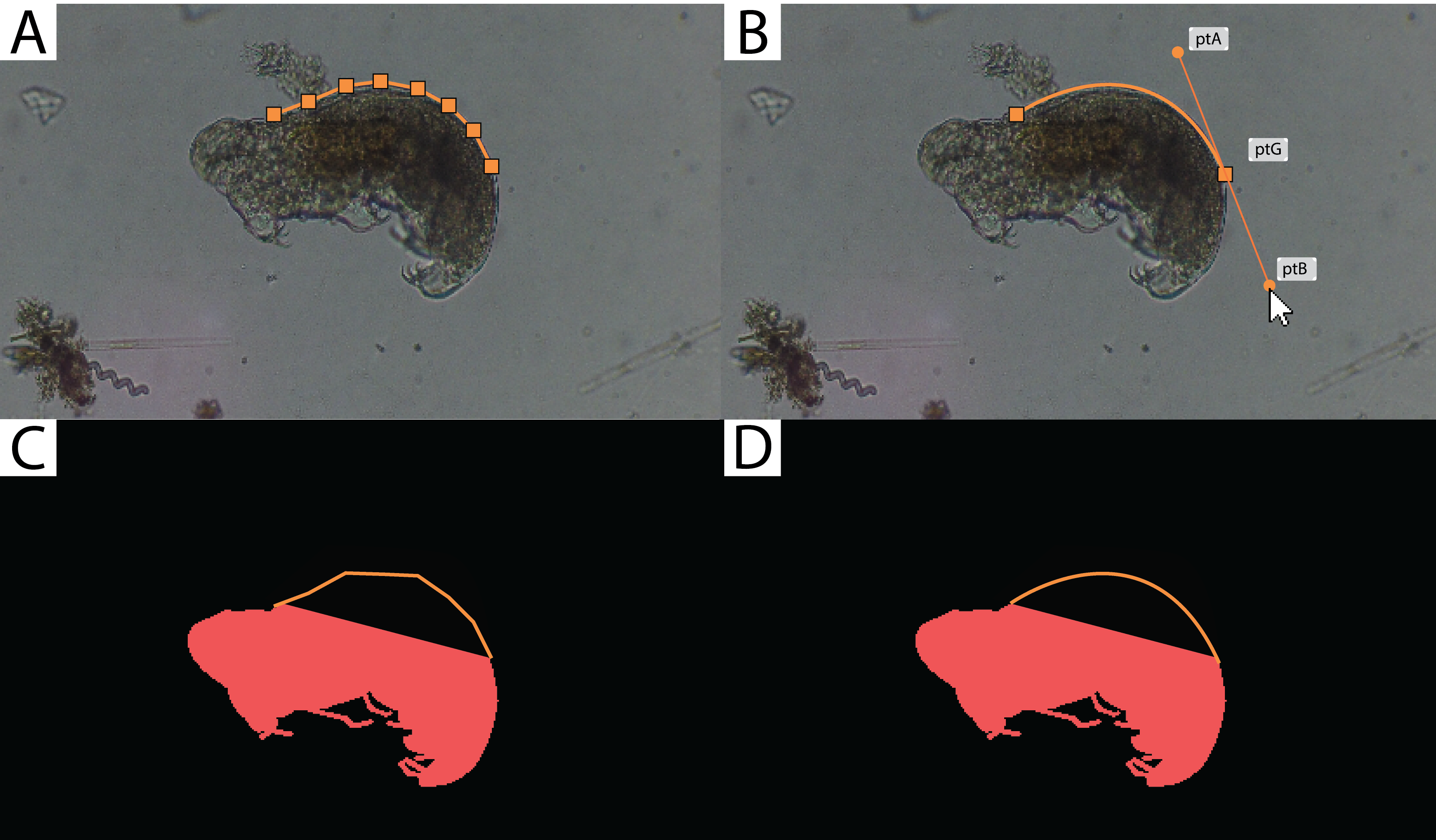}
    \caption{Comparison between polygon (panels A and C) and Bézier-splines (panels B and D) in the delimitation of boundaries of a tardigrade's body. One Bézier unit is defined by three points. \texttt{ptA} and \texttt{ptB} constitute the start and end of the unit, whereas \texttt{ptG} serves as a point of gravity to pull the line away from the user's cursor, forcing it to become a curve.}
    \label{fig:BézierSplines}
\end{figure}

A central feature of LivePyxel is its support for Bézier splines \cite{rogers2000introduction}; a mathematical representation of curves that originated in computer graphics and has become an industry standard for precision editing. Unlike traditional polygon tools, which approximate shapes using straight-line segments and are better suited for rigid structures like crystals, buildings, or mechanical parts, Bézier splines allow for smooth, continuous contours; see Fig.~\ref{fig:BézierSplines}. This makes them ideal for segmenting organic, curved shapes typically encountered in biological datasets, such as cells, tissues, or protozoa.

Each Bézier unit is defined by three control points: two end points and a central handle that determines curvature; see Fig.~\ref{fig:BézierSplines}-B. By connecting multiple units, users can construct complex outlines with high fidelity to natural forms. This modular structure offers fine control over both curvature and sharp transitions. 
The advantage of splines for delimiting contours is illustrated by SplineDist \cite{mandal2021splinedist}, which extends over the popular StarDist framework by modeling objects as planar parametric spline curves, allowing more flexible and smooth segmentation boundaries and solving issues with non-convex geometries. Moreover, the use of splines in the preparation of annotated data for segmentation tasks has been documented in clinical applications \citep{huellebrand2022collaborative, materka2024using}.
In biomedical imaging, the use of splines enables users to trace smooth cell membranes with highly organic contours and tightly coiled or angular biological features, something that would require excessive effort and precision with polygon tools.
The ability to produce accurate, pixel-level masks with fewer interactions not only improves annotation speed but also reduces user fatigue and annotation bias. As a result, Bézier splines in LivePyxel improve both the efficiency and the quality of segmentation in tasks where precision is critical, such as microscopy, radiography, and digital morphology.

\subsection{Binary Masks and Multi-Layered Structure}
Another important feature provided by LivePyxel is the generation and manipulation of binary masks, which enable semi-automated annotation in highly controlled imaging environments. LivePyxel generates binary masks by applying pixel-level operations using OpenCV throughout the frame, then assigning a value of $0$ or $1$ to every pixel before creating an annotation mask. Using this technique, the user can isolate regions of high contrast against a uniform background, such as a white or black field. This proves to be especially useful in biomedical applications involving solid and organic objects \cite{chen2023bezierseg,galvez2021nurbs}, and potentially significant in many other fields of biological science, where annotation targets are organic and contour-heavy. We showcase this feature in Section~\ref{sec:snail_shells}.

Binary masks significantly reduce annotation time by providing an initial segmentation that can be manually refined, in contrast to beginning an annotation from scratch. Particularly useful in situations where there are many structures to annotate in a controlled environment with a static background (see Section \ref{sec:snail_shells} targeting an example using snail shells), LivePyxel provides a way to critically boost the gathering of image/mask pair data. The generated dataset can further be augmented or `engineered' to generate a much larger dataset (Fig.~\ref{fig:dataEng}).

Finally, binary masks can also be used in combination with vector-based tools such as Bézier splines. Once a region is roughly defined using thresholding, the user can trace or refine its boundaries using spline functions to achieve pixel-level precision. This hybrid workflow enhances segmentation quality while minimizing manual effort, offering a powerful solution for datasets where both speed and anatomical accuracy are crucial.

\begin{figure}[h]
  \centering
    \includegraphics[width=0.95\columnwidth]{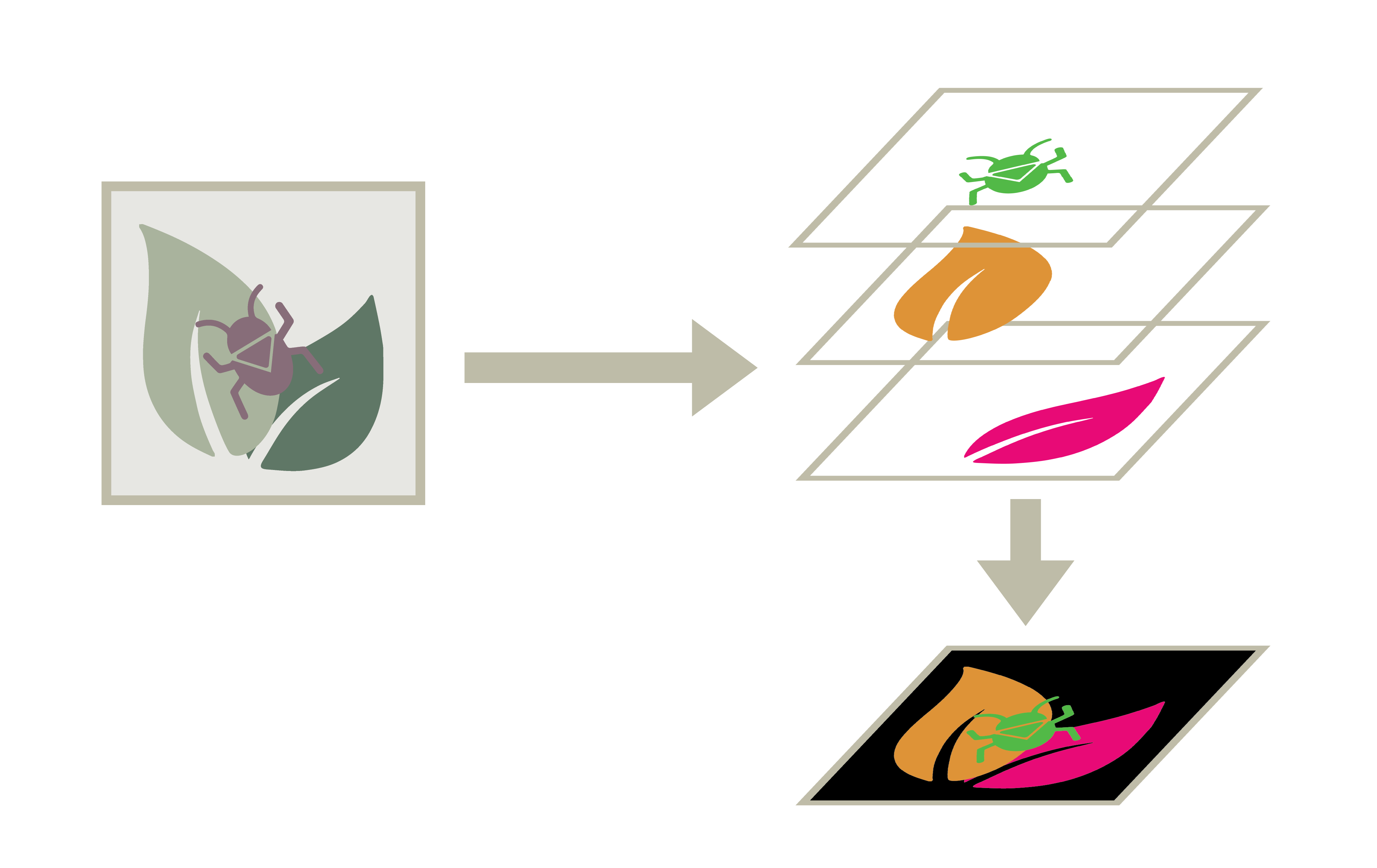}
    \caption{Illustration of the multi-layer annotation process. Each annotation is stored as a separate layer within a stack, which is subsequently merged into a single raster image from bottom to top upon saving.}
    \label{fig:multilayer}
\end{figure}

LivePyxel employs a stack-based compositing system, in which each annotation exists on an independent layer logically stacked from bottom to top. This layered architecture facilitates non-destructive editing: individual segments or anatomical regions can be added, modified, or removed without affecting neighboring annotations; see Fig.~\ref{fig:multilayer}. Each layer retains its own shape, color, and mask information, granting users fine-grained control over the annotation workflow. When the \textit{Annotate} button is pressed, these layers are rasterized and composited in order, from bottom to top, into a single merged mask, encoding the labels in the form of RGB colors and overwriting every pixel with the top color (ignoring black). This final image can be saved or exported in formats compatible with most deep learning frameworks. 

Importantly, this integrated pipeline also facilitates instance segmentation tasks, where models must not only classify each pixel but also associate it with a specific object instance. By storing each layer separately, LivePyxel can export instance-specific masks, with each layer representing a distinct object. This capability enables seamless integration with training workflows for models such as Mask R-CNN and SAM.

\subsection{Webcam integration} 
LivePyxel, through the OpenCV library, integrates the webcam feed into the GUI.
The program begins by searching for up to four camera devices connected to the computer, displaying the first device found. This multicamera feature is useful in situations where the user has multiple video-device stations connected to the application and may require flexibility in accessing different cameras in controlled environments.
Once loaded, webcam images are rendered onto the canvas, allowing users to perform annotations either in on-site during live streaming or on captured frames.
The user concludes an annotation by clicking on the \texttt{Annotate} button. The annotation will be stored in the folder specified by the user, in the form of an RGB image in the native webcam's resolution, and a matching mask with the pixel annotations encoded by the colors chosen by the user.
LivePyxel is capable of reading most formats supported by OpenCV, including \texttt{.jpg}, \texttt{.png}, \texttt{.tiff}, and \texttt{.bmp}, and the annotated images are saved in \texttt{.png} format. The identity and color of each labeled class are stored in a \texttt{.json} file for its readability and its common use in a wide range of programming and app development applications.

\subsection{Tablet integration} 
The use of drawing tablets in combination with graphics editors is a desirable practice as it provides notable advantages over the use of a computer mouse. Tablet integration improves annotations by reducing the time required to delimit the contour of objects without compromising precision. The libraries that integrate the core functionality of the program,  OpenCV and pyQt,  contain highly optimized compatibility with computer mouse-like devices, resulting in a smooth ``plug-and-play" integration with LivePyxel. The ``Free Hand" is a specific tool designed to improve the user experience with drawing tablets. It allows the user to automatically generate new vertices in a polygon by calculating the distance between the last vertex and the position of the tablet's stylus, adding a new vertex when the distance is greater than a reference threshold. This behavior allows the user to focus entirely on the hand movement, tracking the contour. This functionality was essential for creating the dataset used to train a vision model to segment eight different microorganisms in Section~\ref{sec:water_tank}, using an Artist Display 15.6 Pro tablet.



\section{Examples}\label{sec:examples}
Here, we present two distinct examples demonstrating the use of LivePyxel for labeling training data in pixel-level vision tasks. In Section~\ref{sec:water_tank}, we showcase its application in a segmentation task involving eight different microorganisms, highlighting manual data annotation using Bézier splines and other built-in tools. In Section~\ref{sec:snail_shells}, we describe the development of a large dataset through semi-automated labeling with binary masks, followed by data augmentation techniques to evaluate LivePyxel’s scalability in dataset creation for training vision models. 
A video introduction to the use of binary masks for the semi-automation of dataset annotations is presented at: 
\href{https://youtu.be/i6wUTPyAq2U}{tutorial link}.
For both examples, the training pipelines and models were implemented using PyTorch \cite{pytorch}.
Finally, in Section \ref{subsec:benchmarkTools}, we compared the labeling accuracy of LivePixel with four other annotation tools.

\subsection{Water Tank: Image Segmentation Task}\label{sec:water_tank}
The presence of certain microorganisms in the environment often serves as an indirect indicator of ecosystem health \cite{cairns1993proposed}. These organisms, commonly referred to as indicator species (IS), are highly sensitive to chemical and physical changes in their environment, and their use has become widespread in environmental assessment studies \cite{siddig2016ecologists}. Some datasets and data management digital infrastructures have been published in recent years \cite{li2021emds, li2020developing, goldberg2005open}, highlighting the growing importance of automation in the identification and monitoring of these species communities. However, the ephemeral nature of these communities poses significant challenges in collecting and processing annotations for vision-model tasks at a rate comparable to the speed at which such communities change.
To demonstrate the capabilities of LivePyxel, we curated a high-quality dataset capturing the biodiversity of water samples collected from an urban park in Toronto, Canada, in April 2025. We acquired 1,250 image-mask pairs from a water tank, belonging to eight categories of microorganisms (see Supplementary Material for more information).

We trained a U-Net model \cite{ronneberger2015u} for image segmentation across eight categories, initializing it with VGG-19 pretrained weights \cite{simonyan2014very}. Each image in the dataset had an original resolution of 720$\times$480 pixels. Due to the dataset’s limited size, certain categories were underrepresented (see Fig.~S.2 in the Supplementary Material). To mitigate this imbalance, we applied categorical weighting to the cross-entropy loss function and employed data augmentation techniques. Additional architectural and training details are provided in the Supplementary Material.
Training was carried out over 131 epochs using two NVIDIA V100 16 GB GPUs, with a total training time of approximately 8 hours. For evaluation, we reserved an independent test set of 200 images that were excluded from both training and validation.

\begin{figure}[h!]
  \centering
  \includegraphics[scale=0.4]{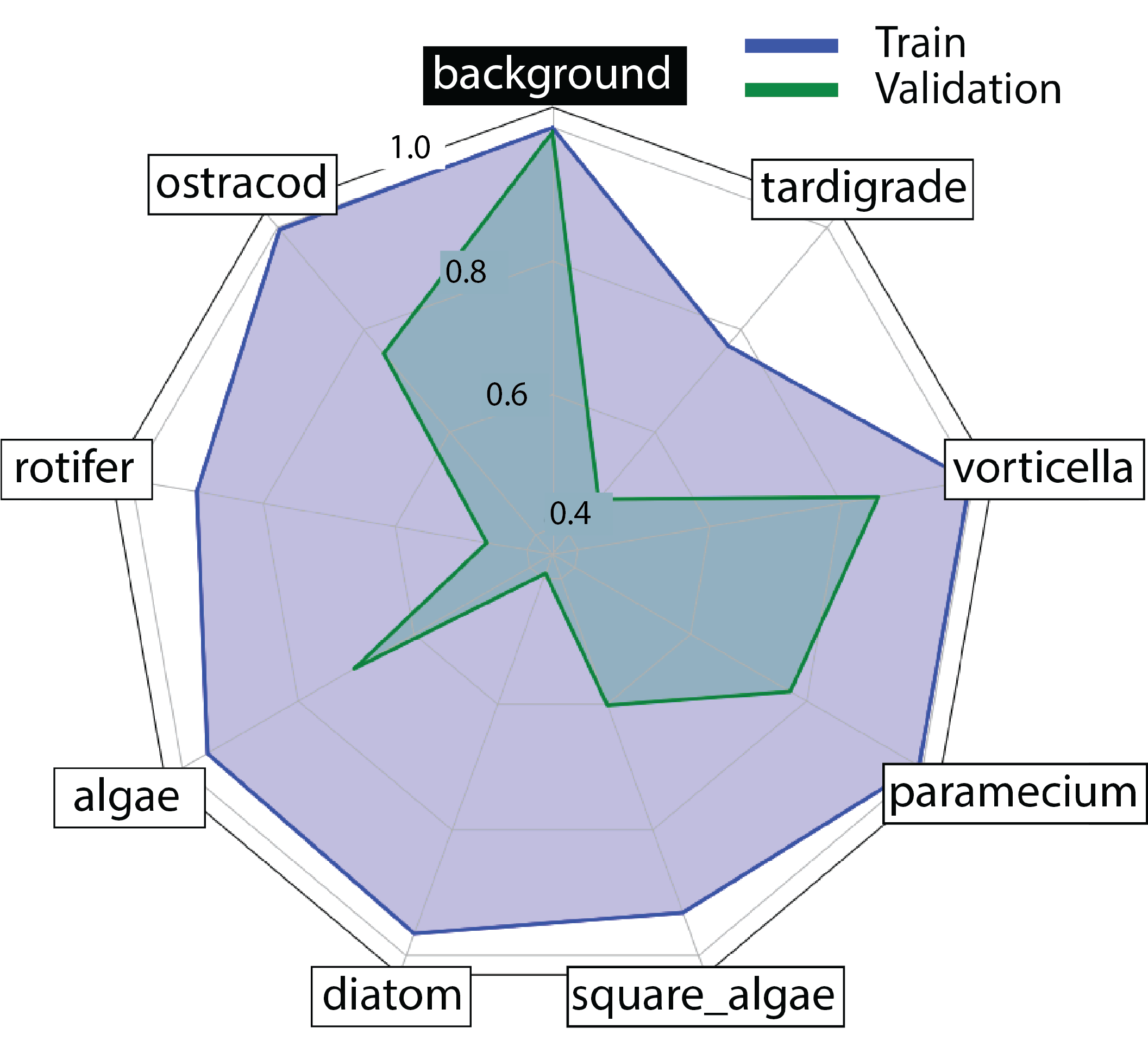}
  \caption{The F1 scores achieved by a U-Net model, highlighting the performance across the eight different microorganisms and the background. 
  The U-Net was initialized with the VGG-19 weights. For more details regarding the training, see the main text. 
  }
  \label{fig:F1RadarEM}
\end{figure}

Fig.~\ref{fig:F1RadarEM} presents the final F1 scores of the trained U-Net model, which performed well in identifying the most abundant classes despite the limited dataset size. These scores reflect consistent accuracy for dominant taxa, but also highlight the challenges of detecting rare or visually ambiguous classes such as tardigrades, diatoms, and rotifers, an issue aligned with the class imbalance of the dataset, where these taxa collectively account for only about 10\% of samples (see Fig.~S.2 in the Supplementary Material). As shown in Fig.~\ref{fig:predictionsMadeEM}, the model also struggled to detect specimens that exhibit transparent bodies, highly variable morphologies, and poor representation in training data, such as \textit{Vorticella} sp., resulting in incorrect contour recognition.

\begin{figure}[h!]
  \centering
  \includegraphics[width=\linewidth]{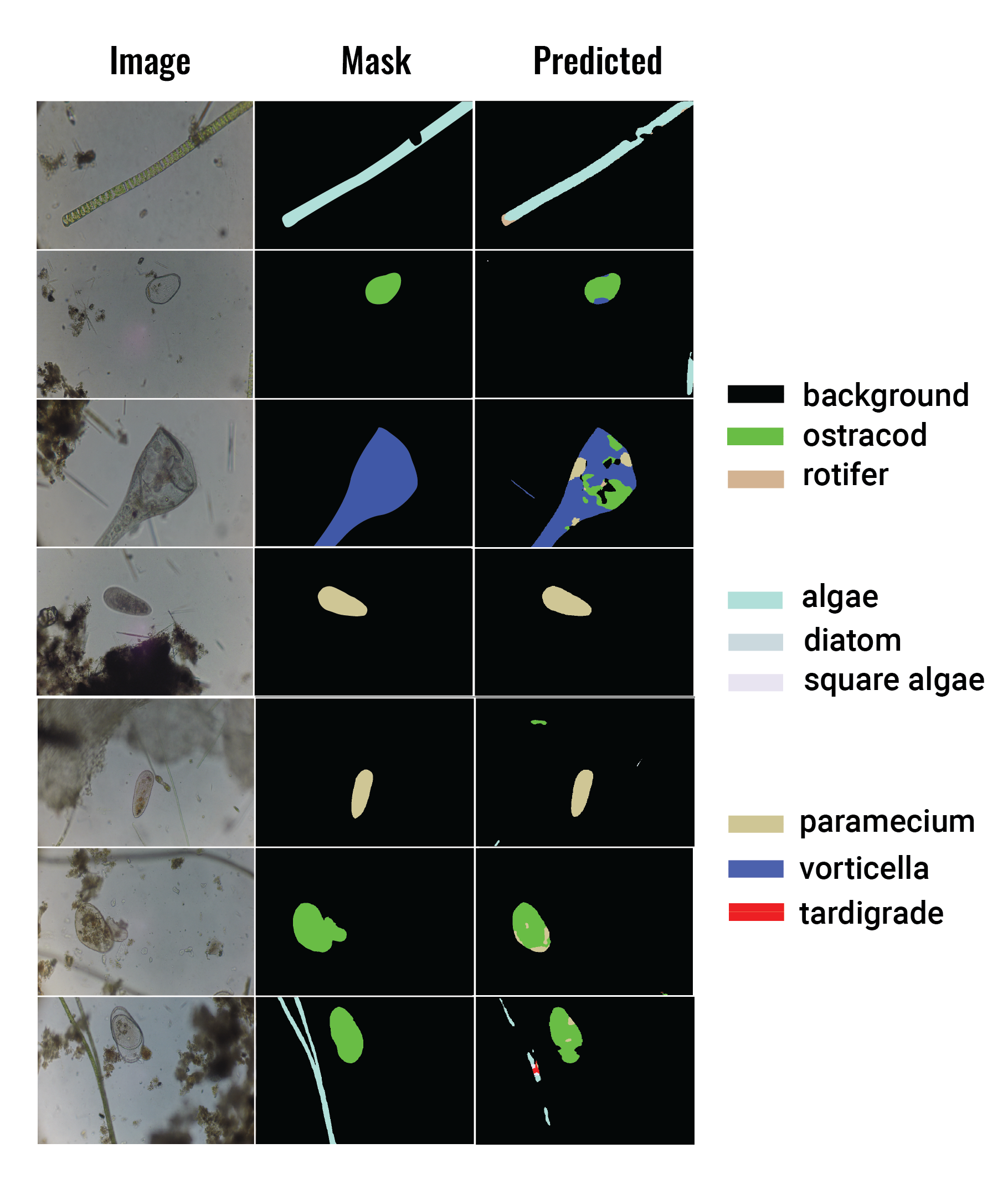}
  \caption{Example predictions from the trained U-Net model on the Water Tank (EM) dataset. Each row shows the original microscopy image (left), the corresponding ground truth mask (middle), and the model’s predicted segmentation (right). The legend on the right indicates the color coding for each class.}

  \label{fig:predictionsMadeEM}
\end{figure}

The Environmental Microorganisms dataset served as an example of the capacity of LivePyxel to rapidly produce quality annotations of a rapidly changing community of species. The U-Net attained reliable F1 scores for the most abundant classes, but performance degraded for rare, transparent, or morphologically variable organisms (e.g., \textit{Vorticella} sp., diatoms, rotifers), reflecting the dataset’s class imbalance. These outcomes underscore two complementary needs for ecological vision models deployed in dynamic microbial communities: (i) the importance of capturing morphological variability; and (ii) the use of training strategies that are robust to scarcity and ambiguous boundaries, including categorical weighting in the cross-entropy loss to account for rare classes, data augmentation and the potential in using data engineering techniques to boost the capacity of the network to recognize rare classes. 

\subsection{Snail Shells: Data engineering with binary masks}\label{sec:snail_shells}
Creating training datasets for semantic segmentation is notoriously labor-intensive. Precise pixel-level annotations are required, and this process becomes especially difficult for images containing many small and round objects, where annotators must painstakingly outline each instance. For example, producing 1,000 segmentation masks for a benchmark like MS COCO took about 22 \cite{ghiasi2021simple} hours, highlighting the high cost of manual labeling. The performance of most vision models improves significantly with the size of the training dataset. This creates a strong incentive to explore automated and semi-automated annotation methods and advanced data augmentation to expand datasets without proportional human effort \cite{ghiasi2021simple}.

In certain scenarios, a dataset can be constructed under highly controlled lighting and background conditions to semi-automate the annotation process. For example, using a uniform background (e.g., a solid white or green backdrop) and exploiting simple threshold-based image operations to obtain binary masks for the foreground objects. In a controlled setup where the background is a known uniform color (such as white), images can be converted to grayscale (or other color space) and apply a threshold that classifies each pixel as background ($0$) or object ($1$) based on its intensity or hue \cite{garcia2021image}. This effectively binarizes the image, separating the objects from the backdrop. For example, researchers have captured objects on a plain white background under consistent lighting, which reduces the complexity of background removal and allows easy discrimination of the object by simple intensity thresholding when dealing with organic shapes \cite{mohana2015automatic}. In fields like medical imaging, placing surgical instruments in front of a green screen has been used to automatically extract tool masks – the monochromatic green background makes it easy to isolate the instrument by hue thresholding \cite{garcia2021image}. In such controlled conditions, the threshold can be tuned so that any pixel brighter (or darker) than a set value is labeled as background, while the rest are labeled as foreground (object). This semi-automated mask generation dramatically speeds up dataset creation, since the bulk of the annotation is handled by an algorithm (with minimal human correction for any errors).

We used LivePyxel to test its capabilities in the automation and generation of image/mask pairs in a highly controlled environment using an assortment of snail shells purchased at a dollar store in Toronto, Canada. The dataset is composed of 4 different classes, believed to correspond to different species of mollusks (see Fig.~S.3 in the Supplementary Material).

\begin{figure}[h]
  \centering
    \includegraphics[width=\columnwidth]{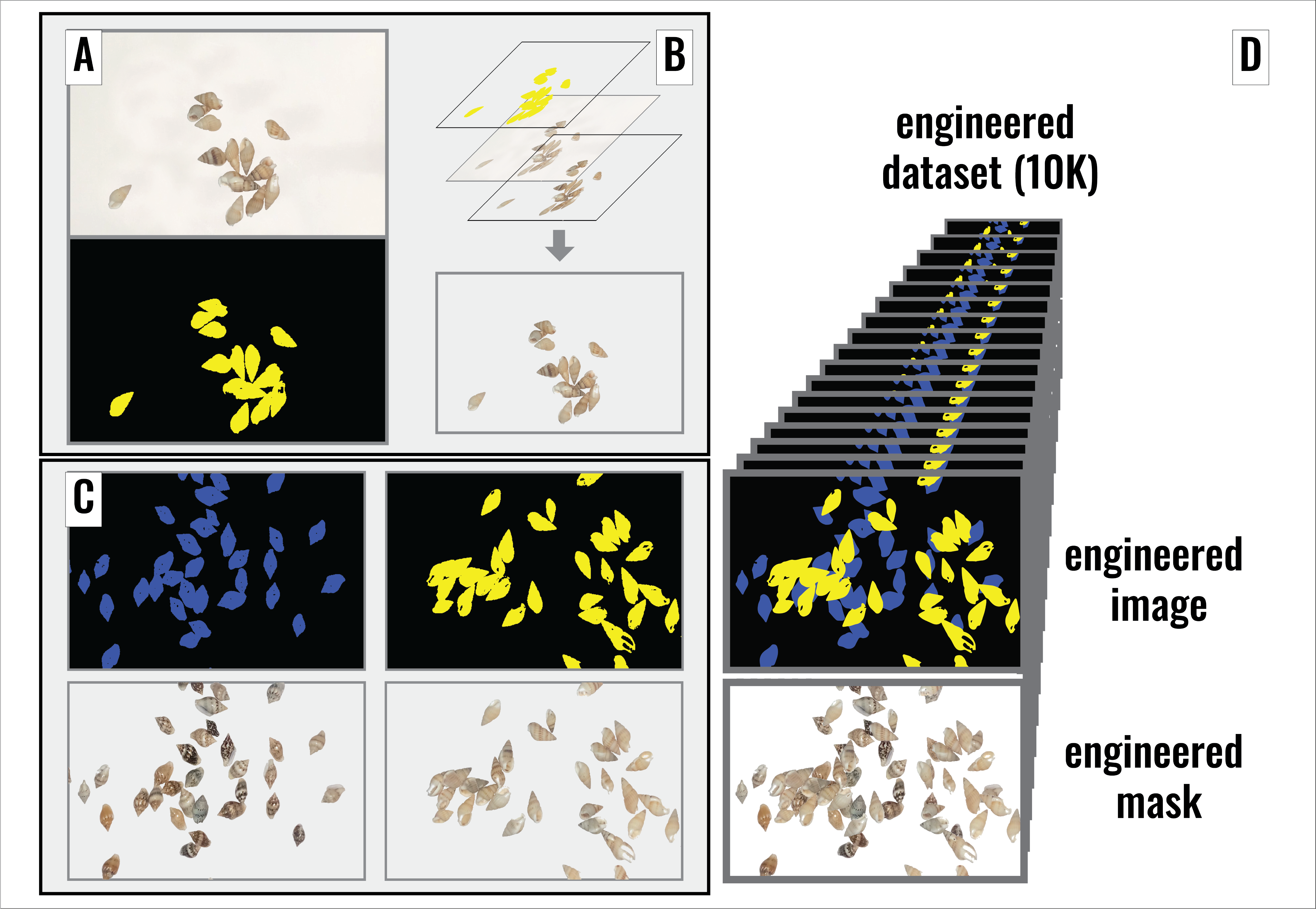}
    \caption{Pipeline for Engineered Semantic Segmentation Dataset Creation. (Panel A) The binary mask tool from LivePyxel captures an image/mask annotation of shells belonging to the same class. (Panel B) Masks are overlapped on top of images to retrieve transparencies. (Panel C) Transparencies are randomly stacked together, forming a new composite image on top of a background. (Panel D) We quickly engineered a dataset consisting of 10,000 image/mask pairs.}
    \label{fig:dataEng}
\end{figure}

The original training data set was collected by placing a large number of shells from the same category at the same time, using the binary mask tool to discriminate the background from the shells, and assigning a color to the final annotation (Fig.~\ref{fig:dataEng}-~A). This resulted in 1,400 individual annotations, each containing only a single class per image. To prepare these data, we used each binary mask to isolate the pixels of the object class within the original image. This yielded a trimmed version of the object with transparency, where all background pixels were removed and only the object pixels remained (Fig.~\ref{fig:dataEng}-~B). These trimmed images were then stored along with their masks.

We implemented a randomized compositing strategy. For each of the 10,000 engineered samples, a base background image was selected, and the class folders were shuffled to introduce variation. A random number of classes were sampled, and for each selected class, a transparent image was randomly chosen and placed on the background (Fig.~\ref{fig:dataEng}-~C). The same transformation, such as flipping or rotation, was applied to both the image cutout and its corresponding mask. This ensured that pixel-level alignment was preserved within an image/mask pair. Multiple instances from different classes were overlaid in succession, resulting in composite scenes that mimic realistic configurations with precise pixel-accurate masks for each object (Fig.~\ref{fig:dataEng}-D).

\begin{figure}[h]
  \centering
  \includegraphics[scale=0.45]{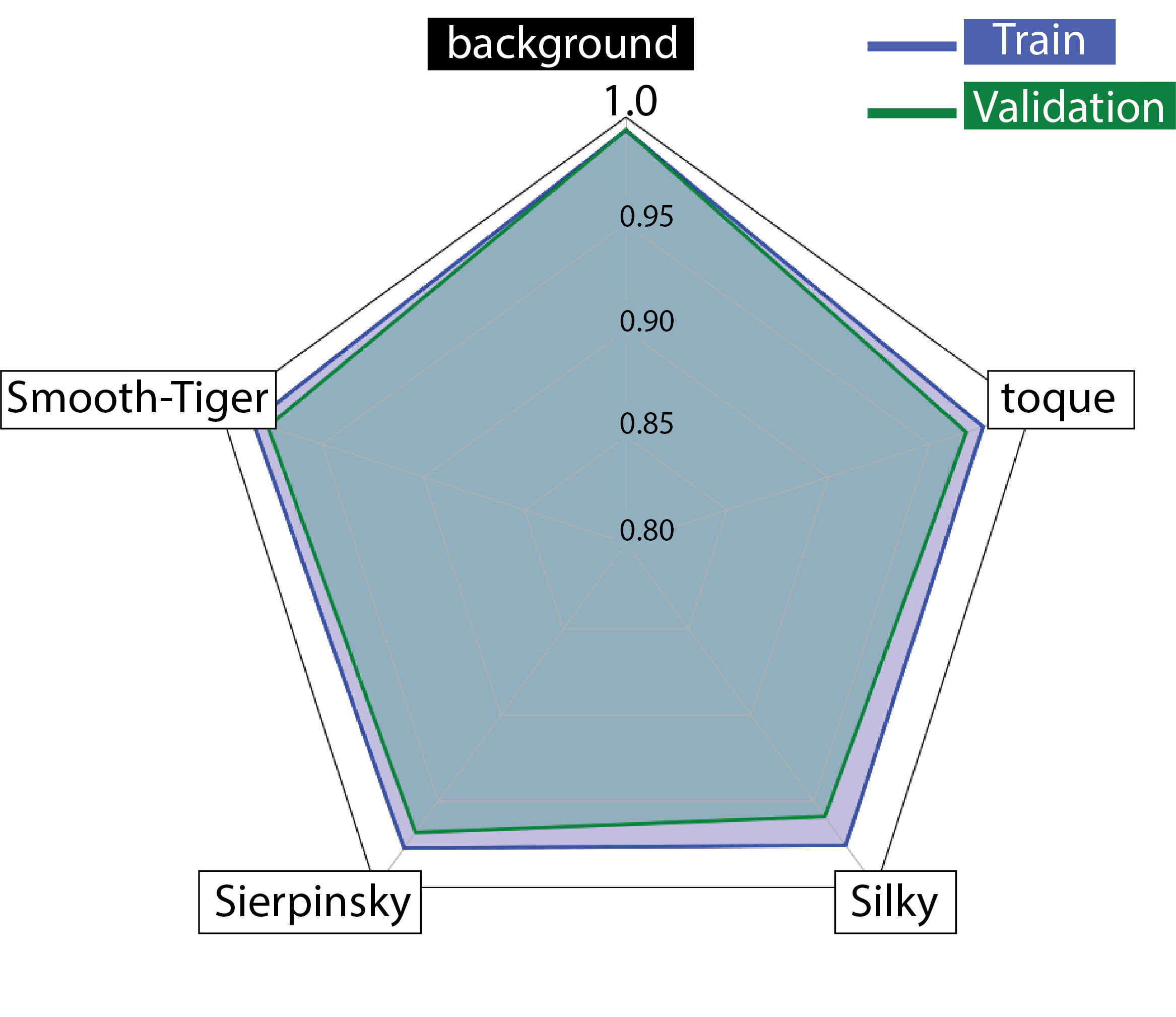}
  \caption{F1 scores achieved by a U-Net model with VGG-19 backbone during training and validation. The plot highlights performance across 5 different classes.}
  \label{fig:radarSnails}
\end{figure}

This pipeline produced a total of 10,000 synthetic image/mask pairs, significantly enriching the dataset and introducing diverse combinations of object instances, orientations, and overlaps. These engineered samples were subsequently used to train the same U-Net segmentation model used for Section~\ref{sec:water_tank}, keeping the image size at 512. The model was trained for 24 hours on 4 NVIDIA A100 GPUs.
In contrast with the water tank dataset (Section~\ref{sec:water_tank}), the F1 scores for the snail shells dataset demonstrated consistently high performance across all classes, with minimal variation between training and validation sets and a very high F1 score across all categories (Fig.~\ref{fig:radarSnails}). The model also exhibited a rapid decline in the loss function, accompanied by a steep rise in F1 scores within the first three epochs (Fig.~S.5 and Fig.~S.6 in Supplemental Material).
Fig.~\ref{fig:SnailsU-NetResults} showcases the predicted masks by the trained U-Net for some of the images in the dataset. 

\begin{figure}[h]
  \centering
  \includegraphics[width=\columnwidth]{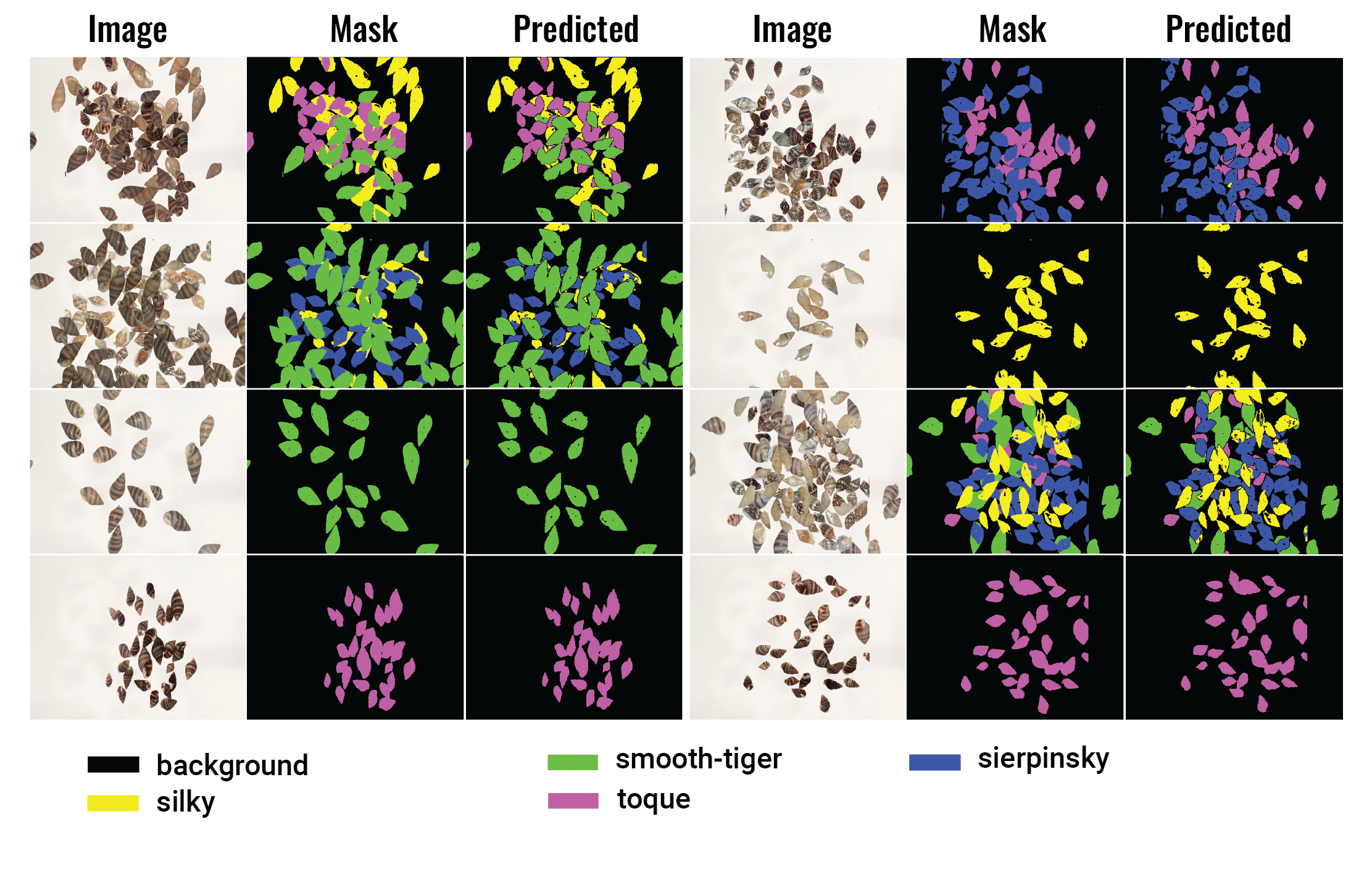}
  \caption{Example predictions from the trained U-Net model on the snail shells dataset. Each set of three columns shows the original image (left), the corresponding ground truth mask (middle), and the model’s predicted segmentation (right). The legend below indicates the color coding for each class.}

  \label{fig:SnailsU-NetResults}
\end{figure}

The snail-shells study demonstrates LivePyxel's capacity to automate the creation of masks under highly controlled setups, yielding fast, stable optimization and uniformly high F1 across classes. Compared to the environmental microscopy setting, the engineered dataset reduces annotation costs in time and user effort. The U-Net architecture, initialized with VGG-19 and trained on 10,000 composite image/mask pairs, converged within a few epochs and exhibited a minimal train–validation gap. These results underscore that with accurate binary masks and controlled backgrounds, segmenting small, round objects becomes straightforward under laboratory conditions and with the use LivePyxel.

\subsection{Comparison between polygon and Bézier-splines}\label{subsec:benchmarkTools}
\begin{figure}[t]
  \centering
  \includegraphics[width=1.\linewidth]{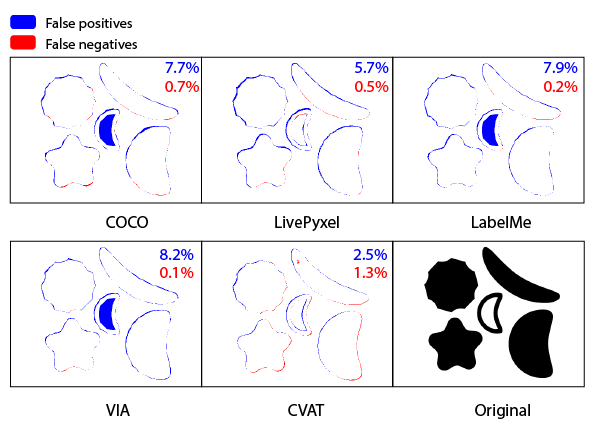}
    \caption{
    Comparison of annotation accuracy across different annotation tools using a reference image (Original) with various levels of curvature complexity. Blue regions represent false positives (areas incorrectly included), while red regions denote false negatives (areas missed). Percentages in each panel indicate the total proportion of false positives (blue) and false negatives (red) for each tool relative to the total number of pixels in the image.
    }
  \label{fig:polygons}
\end{figure}

We evaluated the performance of LivePyxel against several widely used annotation tools. To ensure a fair comparison, a reference image composed of well-defined polygons with varying degrees of curvature complexity was generated (panel labeled "Original"  in Fig.~\ref{fig:polygons}).

Our results show that the accuracy achieved with LivePyxel is comparable to, or slightly higher than, other annotation software packages, while maintaining similar annotation times; see Section 5 in the SM. As highlighted in Section~\ref{sec:snail_shells}, a key advantage of LivePyxel is its capability to perform Boolean operations directly within the mask. This feature, uncommon among existing tools, enables rapid and flexible labeling of complex regions, including those with internal holes like the central object in Fig.~\ref{fig:polygons}.

In terms of performance, LivePyxel exhibited a balanced trade-off between false positives (5.7\%) and false negatives (0.5\%), comparable to other tools (Fig.~\ref{fig:polygons}). 
CVAT achieved the lowest overall error (2.5\% false positives, 1.3\% false negatives) through AI-assisted segmentation, but at the cost of a more complex setup, internet dependency, and non-local data handling. VIA offered the simplest installation (a standalone HTML file), whereas LabelMe required manual dependency management on some systems. COCO Annotator had the most challenging setup, involving Docker and SQL-based database configuration. Among these, only LivePyxel integrates Bézier-spline support, providing smoother boundary representation than polygon-based tools such as VIA, LabelMe, and COCO Annotator.

\section{Summary}
We present LivePyxel, an open-source pixel-level annotation library designed for both on-site image labeling and pre-saved image datasets. Built on OpenCV, LivePyxel efficiently handles image processing tasks and introduces Bézier splines as a key feature, enabling smoother and more precise mask boundaries.
To demonstrate LivePyxel's utility, we provide examples where the labeled images, generated with LivePyxel, are used for training a vision model, completing the annotation-to-inference loop. The datasets for both presented examples were captured using a GoPro camera mounted on a microscope, highlighting LivePyxel's compatibility with diverse imaging setups.
Finally, LivePyxel supports the use of tablet-type devices, making the annotation process more ergonomic.  
In the near future, we aim to implement two additional features, i) zoom integration, allowing users to work on finer details within LivePyxel's central canvas (Fig.~\ref{fig:livepyxel_gui}-E), and ii) inspired by CVAT's AI integration tools, adding a tool capable of loading AI vision models like SAM to help accelerate the annotation process. 
Even in its present form, LivePyxel represents a significant innovation for labeling images, addressing an unmet need for smoother editing tools that could accelerate data generation.

\section*{Author Contributions}
UG wrote the code for LivePyxel, annotated the datasets, and performed all experiments.
JOO and RAVH guided the project. All authors wrote and approved the final version of the manuscript.

\section*{Conflicts of interest}
The authors have no conflicts to disclose.

\section*{Data availability}
LivePyxel and all trained vision models and data presented in this paper are freely available at \url{https://github.com/UGarCil/LivePyxel} and can be installed through PyPI.

\section*{Acknowledgment}
JOO acknowledges the support from NSERC Discovery Grant No. RGPIN-2024-06670.
RAVH acknowledges the support from the Digital Research Alliance of Canada and NSERC Discovery Grant No. RGPIN-2024-06594.



\bibliography{rsc} 
\bibliographystyle{rsc} 

\end{document}


\pagestyle{fancy}
\thispagestyle{plain}
\fancypagestyle{plain}{
\renewcommand{\headrulewidth}{0pt}
}

\makeFNbottom
\makeatletter
\renewcommand\LARGE{\@setfontsize\LARGE{15pt}{17}}
\renewcommand\Large{\@setfontsize\Large{12pt}{14}}
\renewcommand\large{\@setfontsize\large{10pt}{12}}
\renewcommand\footnotesize{\@setfontsize\footnotesize{7pt}{10}}
\renewcommand\scriptsize{\@setfontsize\scriptsize{7pt}{7}}
\makeatother

\renewcommand{\thefootnote}{\fnsymbol{footnote}}
\renewcommand\footnoterule{\vspace*{1pt}%
\color{cream}\hrule width 3.5in height 0.4pt \color{black} \vspace*{5pt}} 
\setcounter{secnumdepth}{5}

\makeatletter 
\renewcommand\@biblabel[1]{#1}            
\renewcommand\@makefntext[1]%
{\noindent\makebox[0pt][r]{\@thefnmark\,}#1}
\makeatother 
\renewcommand{\figurename}{\small{Fig.}~}
\sectionfont{\sffamily\Large}
\subsectionfont{\normalsize}
\subsubsectionfont{\bf}
\setstretch{1.125} 
\setlength{\skip\footins}{0.8cm}
\setlength{\footnotesep}{0.25cm}
\setlength{\jot}{10pt}
\titlespacing*{\section}{0pt}{4pt}{4pt}
\titlespacing*{\subsection}{0pt}{15pt}{1pt}

\fancyfoot{}
\fancyfoot[LO,RE]{\vspace{-7.1pt}\includegraphics[height=9pt]{head_foot/LF}}
\fancyfoot[RO]{\footnotesize{\sffamily{1--\pageref{LastPage} ~\textbar  \hspace{2pt}\thepage}}}
\fancyfoot[LE]{\footnotesize{\sffamily{\thepage~\textbar\hspace{3.45cm} 1--\pageref{LastPage}}}}
\fancyhead{}
\renewcommand{\headrulewidth}{0pt} 
\renewcommand{\footrulewidth}{0pt}
\setlength{\arrayrulewidth}{1pt}
\setlength{\columnsep}{6.5mm}
\setlength\bibsep{1pt}

\makeatletter 
\newlength{\figrulesep} 
\setlength{\figrulesep}{0.5\textfloatsep} 

\newcommand{\topfigrule}{\vspace*{-1pt}%
\noindent{\color{cream}\rule[-\figrulesep]{\columnwidth}{1.5pt}} }

\newcommand{\botfigrule}{\vspace*{-2pt}%
\noindent{\color{cream}\rule[\figrulesep]{\columnwidth}{1.5pt}} }

\newcommand{\dblfigrule}{\vspace*{-1pt}%
\noindent{\color{cream}\rule[-\figrulesep]{\textwidth}{1.5pt}} }

\makeatother

\twocolumn[
  \begin{@twocolumnfalse}
\begin{tabular}{m{0.cm} p{15.5cm} }

 & \noindent\LARGE{Supplemental Material for "\textbf{\texttt{LivePyxel}: Accelerating image annotations with a Python-integrated webcam live streaming}"} \\
 & \vspace{0.3cm} \\

 & \noindent\large{Uriel Garcilazo-Cruz,\textit{$^{a,b,\dag}$}, Joseph O. Okeme\textit{$^{a,b}$}, and Rodrigo A. Vargas-Hernández\textit{$^{a,b,c,\dag}$}} \\

\end{tabular}

 \end{@twocolumnfalse} \vspace{0.6cm}

  ]

\renewcommand*\rmdefault{bch}\normalfont\upshape
\rmfamily
\section*{}
\vspace{-1cm}


\footnotetext{\textit{$^{a}$~Department of Chemistry and Chemical Biology, McMaster University, Hamilton, ON, Canada}}
\footnotetext{\textit{$^{b}$~School of Computational Science and Engineering, McMaster University, Hamilton, ON, Canada}}
\footnotetext{\textit{$^{c}$~Brockhouse Institute for Materials Research, McMaster University, Hamilton, ON, Canada}}
\footnotetext{\dag~garcilau@mcmaster.ca, vargashr@mcmater.ca}

\rmfamily 

The purpose of this supplemental material is to provide additional details about the proposed work in the main draft. Section \ref{sec:water_tank} presents the additional details of the vision model for microscopic organism segmentation, including hardware specifications, model architecture, dataset characteristics, and training performance. Section \ref{sec:snail_shells} complements this with a different application case focused on data engineering using snail shell images, detailing its specialized dataset and training results. Each section is organized to provide: (1) hardware/experimental setup details, (2) model architecture specifications (with Table~\ref{tab:U-Net_architecture} being particularly relevant for technical implementation), (3) dataset composition analysis (including class distributions shown in Figures~\ref{fig:classFreq} and \ref{fig:dataEng}), and (4) quantitative training evaluations (with F1-score trajectories in Figs.~\ref{fig:f1ScoresEM} and \ref{fig:f1ScoresSnails}). 

\section{Dataset preparation}\label{sec:datasetPrep}
All datasets in the study shared the same preprocessing in preparation to enter the vision model. A python script traversed across all image/mask pairs in the dataset, quantifying the relative frequency of pixels for each category. This information was later used to calibrate the network's cross-entropy loss function and make it frequency-dependent. The dataset was prepared using a custom set of scripts that read the \texttt{config.json} file to generate indexes for each label, and mapped it to each of the masks created by LivePyxel, by replacing the color of the mask with the corresponding index, and having the background color of the mask: (0,0,0), as a category. All required scripts that address the dataset, model, and training modules are available in the \href{https://github.com/UGarCil/LivePyxel}{GitHub repository} examples folder.
Future releases of LivePyxel are planned to include a pipeline to import a trained dataset that automates the 
annotation process using models trained by the user.\\

\section{Vision Model Architecture}

All datasets in this study were trained using the same U-Net network. The original dataset was split into training and validation, using 90\% of image/mask pairs for training and 10\% for validation. The network was trained using a custom number of epochs, but using a batch size of 32, a learning rate of $3 \times 10^{-4}$, the AdamW optimizer, and the cross-entropy loss function. All training was performed using the Compute Canada resources located in the Narval cluster, using a single node with up to 4 GPUs connected.

The U-Net's encoder consisted of four \texttt{downsampling} stages (\texttt{downsample1} to \texttt{downsample4}), each comprising convolutional layers (with ReLU activation) and max-pooling. The first two stages used double convolutions (e.g., Conv2d(3→64→64) and Conv2d(64→128→128)), while the deeper stages (\texttt{downsample3}, \texttt{downsample4}) employed quadruple convolutions (e.g., Conv2d(128→256→256→256→256) to capture complex features. The bottleneck (\texttt{bottle\_neck}) expands the channel dimension to 1024 via two convolutions. The decoder (\texttt{upsample1} to \texttt{upsample4}) used transposed convolutions for upsampling, followed by double convolutions that halved the channel dimensions (1024→512→512), with skip connections concatenating encoder features. The final 1$\times$1 convolution (out) reduced the output to 8 channels, corresponding to the segmentation classes, each encoding a unique color; see Table~\ref{tab:label_colors}.
The model contained 26.3 million trainable parameters, with the decoder and bottleneck layers fine-tuned (as indicated by True in the parameter table), while the encoder weights remain fixed (False). This design balanced computational efficiency with multi-scale feature learning, suitable for pixel-wise segmentation tasks.
We used a U-Net architecture \cite{ronneberger2015u}, composed of an encoder and decoder portion and skip connections. The encoder, or downsampling module of the architecture, was initialized with the weights and biases of the VGG-19 architecture \cite{simonyan2014very} (Table~\ref{tab:U-Net_architecture}).

\begin{table}[h]
  \centering
  \begin{tabular}{|c|l|c|}
    \hline
    \textbf{Index} & \textbf{Label} & \textbf{Color (RGB)} \\ \hline
    0 & ostracod & (0, 255, 0) \\ \hline
    1 & rotifer & (211, 179, 145) \\ \hline
    2 & algae & (164, 251, 233) \\ \hline
    3 & diatom & (202, 215, 220) \\ \hline
    4 & square\_algae & (230, 226, 246) \\ \hline
    5 & paramecium & (207, 198, 149) \\ \hline
    6 & vorticella & (23, 54, 255) \\ \hline
    7 & tardigrade & (255, 8, 8) \\ \hline
  \end{tabular}
  \caption{Labels and corresponding colors used in the segmentation task.}
  \label{tab:label_colors}
\end{table}

\begin{table}[h]
  \centering
  \begin{tabular}{|l|l|c|}
    \hline
    \textbf{Layer} & \textbf{Configuration} & \textbf{Trainable} \\ \hline
    downsample1    & \begin{tabular}[t]{@{}l@{}} 
                      Conv2d(3$\rightarrow$64$\rightarrow$64) \\ 
                      MaxPool2d
                    \end{tabular} & False \\[2ex] \hline
    downsample2    & \begin{tabular}[t]{@{}l@{}} 
                      Conv2d(64$\rightarrow$128$\rightarrow$128) \\ 
                      MaxPool2d
                    \end{tabular} & False \\[2ex] \hline
    downsample3    & \begin{tabular}[t]{@{}l@{}} 
                      Conv2d(128$\rightarrow$256$\times$4) \\ 
                      MaxPool2d
                    \end{tabular} & False \\[2ex] \hline
    downsample4    & \begin{tabular}[t]{@{}l@{}} 
                      Conv2d(256$\rightarrow$512$\times$4) \\ 
                      MaxPool2d
                    \end{tabular} & False \\[2ex] \hline
    bottle\_neck   & \begin{tabular}[t]{@{}l@{}} 
                      Conv2d(512$\rightarrow$1024$\rightarrow$1024) \\ 
                      ReLU(inplace=True)
                    \end{tabular} & True \\[2ex] \hline
    upsample1      & \begin{tabular}[t]{@{}l@{}} 
                      ConvTranspose2d(1024$\rightarrow$512) \\ 
                      Conv2d(1024$\rightarrow$512$\times$2)
                    \end{tabular} & True \\[2ex] \hline
    upsample2      & \begin{tabular}[t]{@{}l@{}} 
                      ConvTranspose2d(512$\rightarrow$256) \\ 
                      Conv2d(512$\rightarrow$256$\times$2)
                    \end{tabular} & True \\[2ex] \hline
    upsample3      & \begin{tabular}[t]{@{}l@{}} 
                      ConvTranspose2d(256$\rightarrow$128) \\ 
                      Conv2d(256$\rightarrow$128$\times$2)
                    \end{tabular} & True \\[2ex] \hline
    upsample4      & \begin{tabular}[t]{@{}l@{}} 
                      ConvTranspose2d(128$\rightarrow$64) \\ 
                      Conv2d(128$\rightarrow$64$\times$2)
                    \end{tabular} & True \\[2ex] \hline
    out            & Conv2d(64$\rightarrow$8, kernel=1$\times$1) & True \\[2ex] \hline
  \end{tabular}
  \caption{U-Net architecture. Encoder layers (fixed weights) downsample via convolutions and max-pooling; decoder layers (trainable) upsample with transposed convolutions and skip connections. ReLU activations follow all convolutions except the final layer. Total parameters: 26.3M.}
  \label{tab:U-Net_architecture}
\end{table}

\section{Examples of Usage: Water Tank} \label{sec:water_tank}
In this section, we provide the necessary information for the segmentation task presented in Section~3.1 from the main draft.
\subsection{Hardware Setup}
\begin{figure}[h]
  \centering
    \includegraphics[width=\columnwidth]{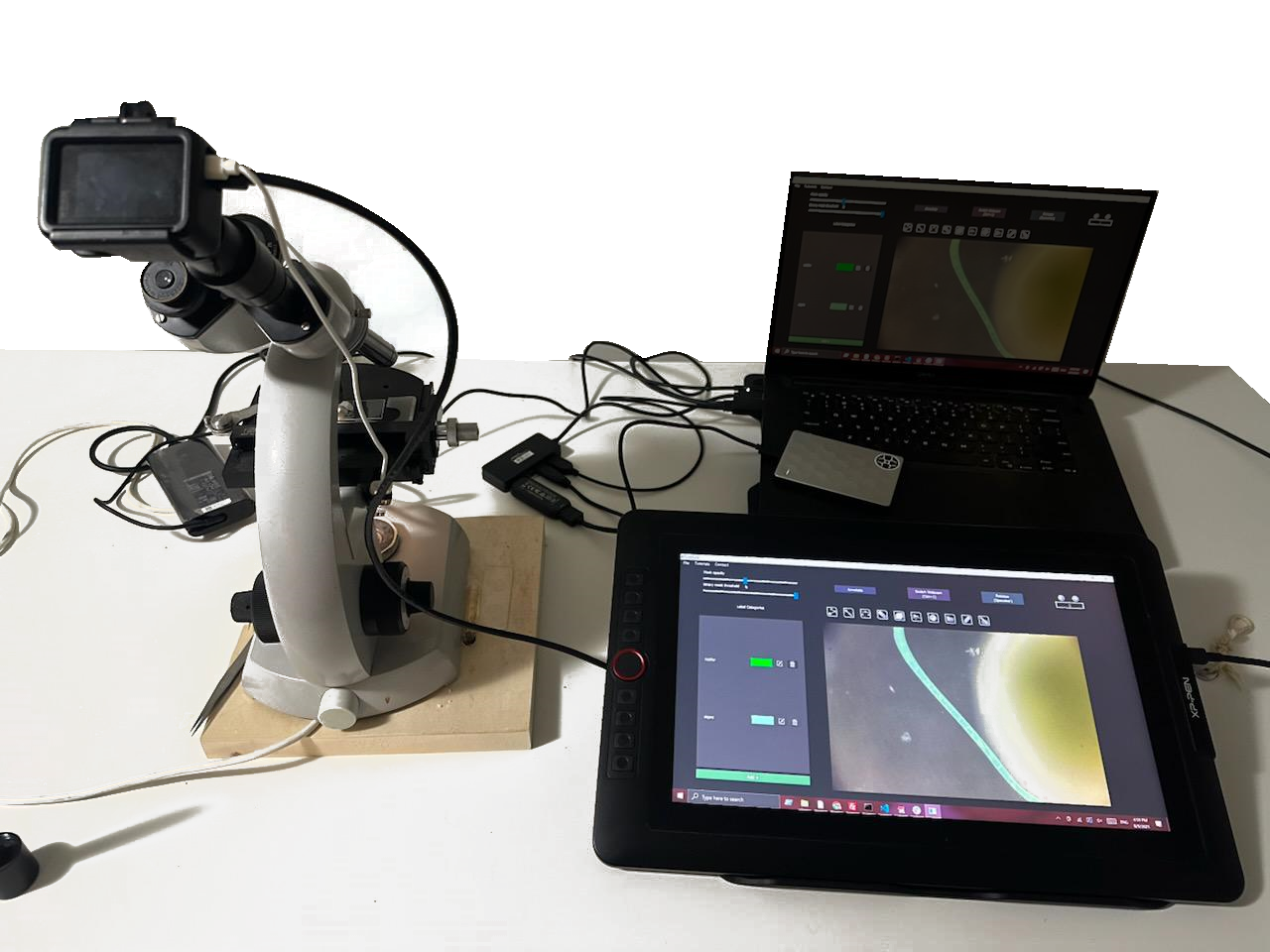}
    \caption{Custom camera setup for deploying LivePyxel. The configuration includes a microscope equipped with a mounted digital camera, connected to a laptop and a pen display tablet for real-time image acquisition and analysis.}
    \label{fig:hardware}
\end{figure}

The imaging setup consisted of a custom configuration using a Zeiss compound microscope (model X-100) with a 10x achromatic objective lens (NA 0.25) and bright-field (Koehler) illumination. Images were captured using a modified GoPro camera equipped with an M12 CCTV lens, affixed to a standard 10x wide-angle eyepiece. A video capture card streamed the camera output as a webcam, ensuring compatibility with OpenCV. A drawing tablet model XP-PEN 15.6 Pro was used to assist with annotations, by duplicating the display and projecting LivePyxel as a secondary screen. Fig.~\ref{fig:hardware} depicts the complete imaging setup.

\subsection{Dataset Information}

\begin{figure}[h]
  \centering
    \includegraphics[scale=0.135]{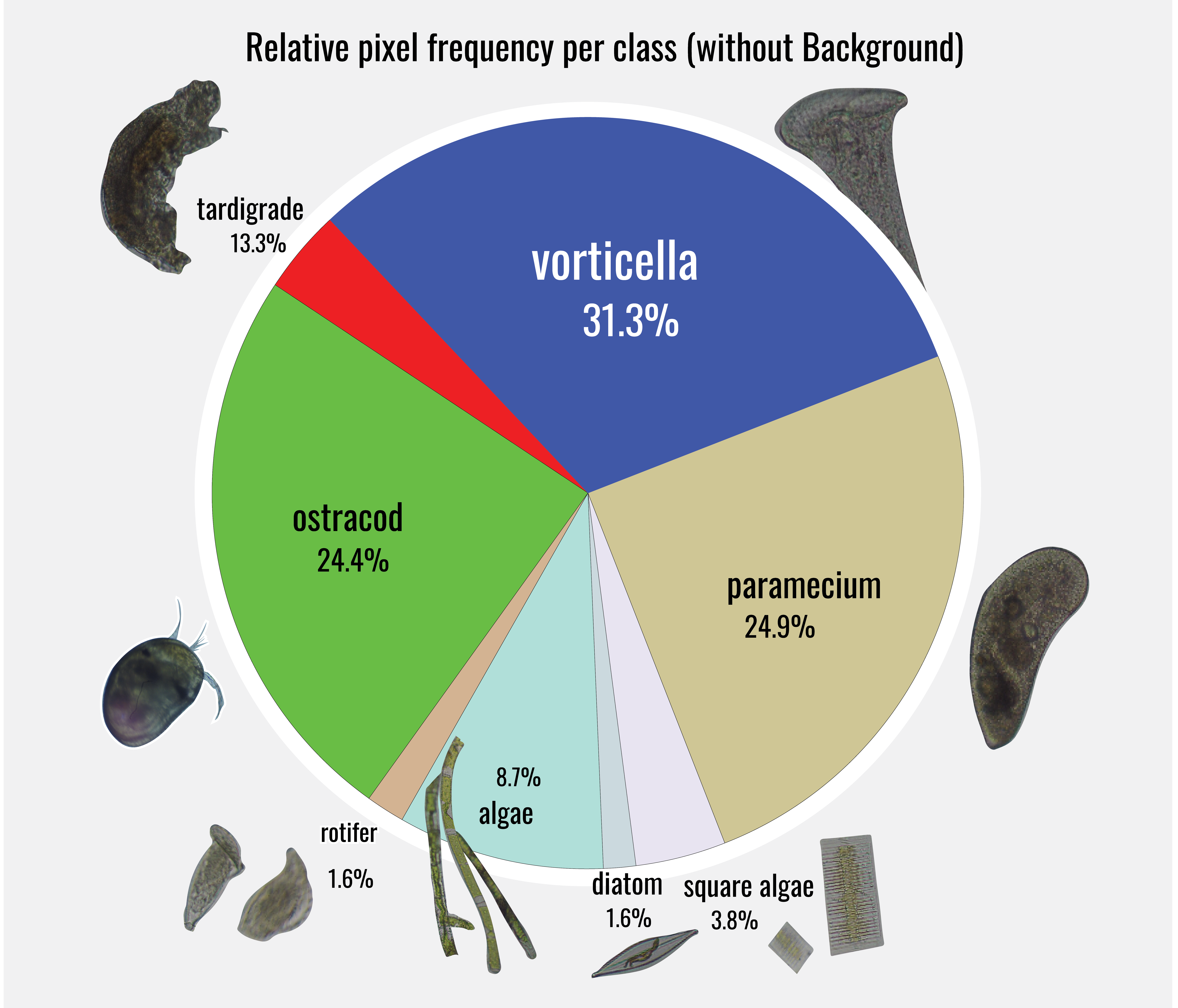}
    \caption{Relative pixel frequency distribution for each annotated class in the EM dataset (N = 1,250 images), excluding background pixels (RGB 0,0,0). Percentages were computed from annotation masks after normalization, showing the proportional representation of \textit{Vorticella}, \textit{Ostracod}, \textit{Paramecium}, and other taxa.}

    \label{fig:classFreq}
\end{figure}

The dataset was composed of 1.25K image/mask pairs, and included a wide asymmetric distribution of categories, as seen in the relative frequency of such classes Fig.~\ref{fig:classFreq}). The three main classes in the dataset were samples of paramecia, vorticella, and green algae, which constitute 79\% of the number of pixels in pixels in any category other than background.

\subsection{Model Architecture}
\label{sec:water_tank_model_architecture}

The water tank dataset was trained on 1,250 images, each with an original resolution of 720$\times$480 pixels. Training was carried out over 131 epochs, with an additional set of 200 images reserved for testing and detection, none of which were used during training or validation. All training was performed using two NVIDIA V100 16 GB GPUs, totaling 8 hours of computation.

Fig.~\ref{fig:trainloss_EM} illustrates the model's performance by plotting the training and validation losses. The plot clearly shows a sharp decline in both training and validation loss from epoch 0 to 20, followed by a gradual increase in validation loss afterward. This pattern suggests the model began overfitting, learning noise from the training data rather than generalizing effectively.

\subsection{Training}

\begin{figure}[h]
  \centering
    \includegraphics[width=\linewidth]{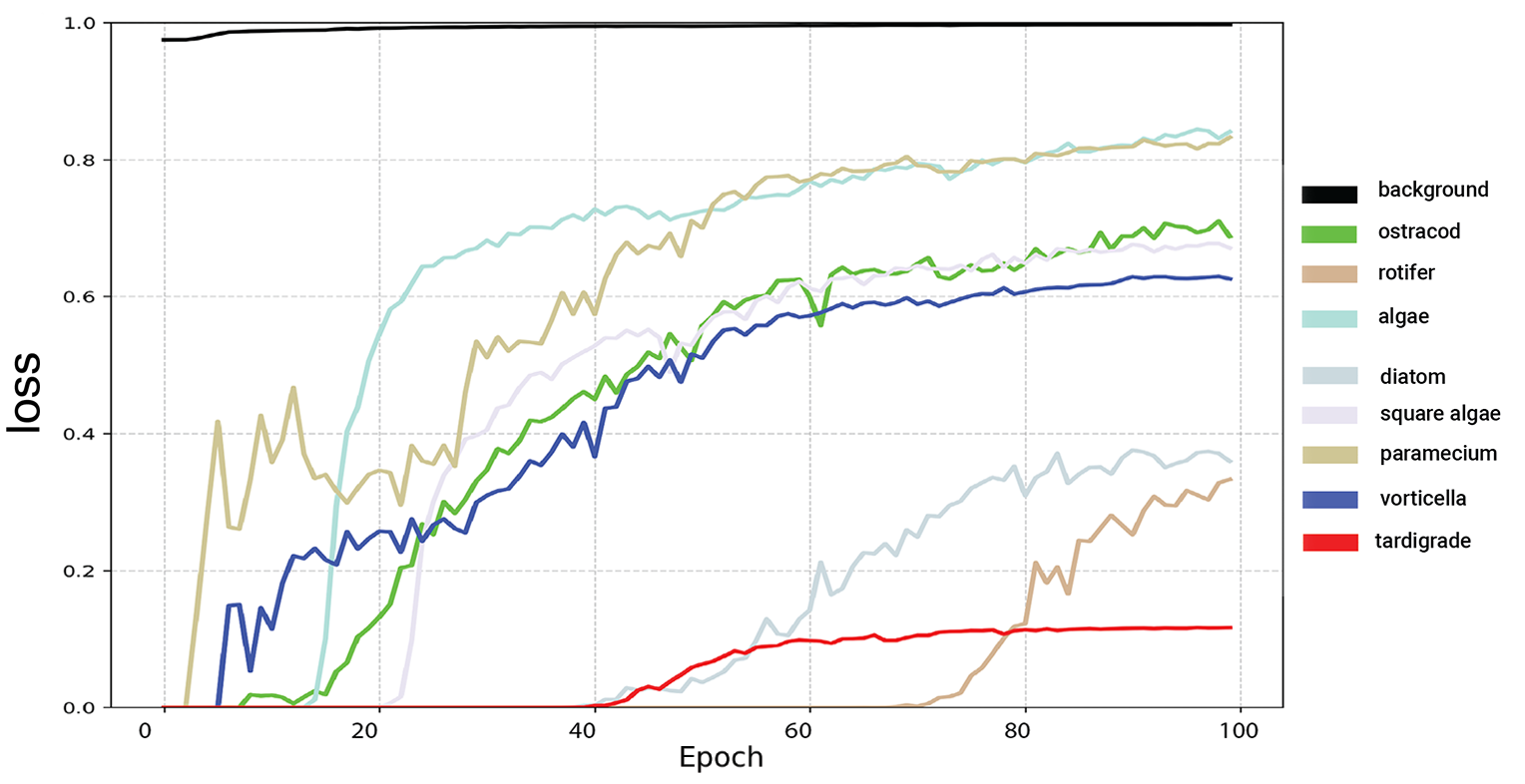}
    \caption{F1 score progression during U-Net training for the nine annotated categories in the EM dataset. The model was initialized with weights from a pretrained VGG19 network. Each curve represents the per-class F1 score over 100 training epochs.}

    \label{fig:f1ScoresEM}
\end{figure}

\begin{figure}[h]
  \centering
    \includegraphics[width=\linewidth]{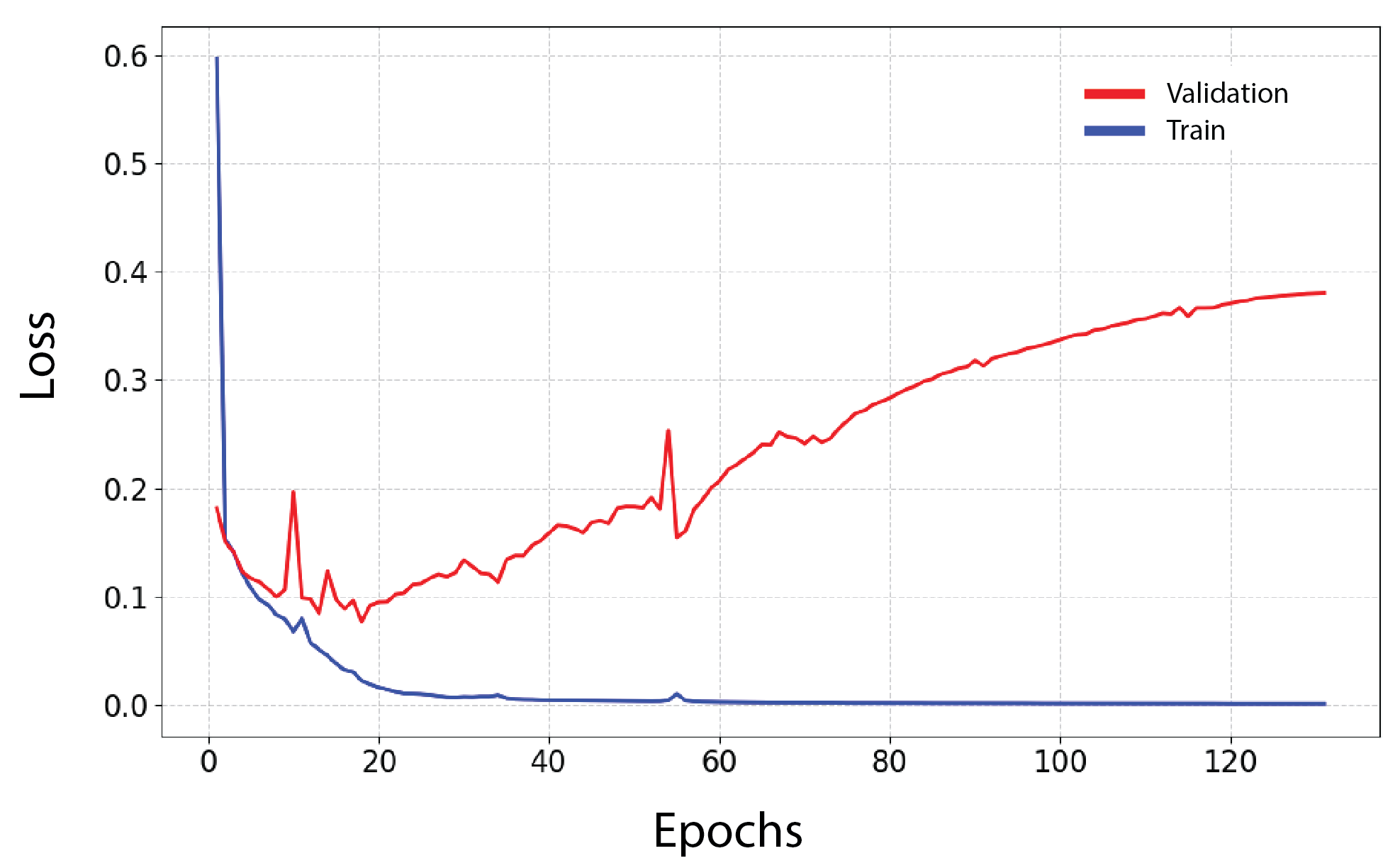}
    \caption{Cross-entropy loss curves for the training and validation sets during U-Net model training on the water tank dataset. The rapid decrease in both losses during the initial epochs is followed by a divergence after approximately epoch 20, indicating the beginning of overfitting.}

    \label{fig:trainloss_EM}
\end{figure}

In addition to the pattern seen in the loss function, the performance of the model can be seen in Fig.~S\ref{fig:f1ScoresEM} with the F1 scores of every class, which rapidly increases in the early training stages and begins to plateau around epoch 20. While the F1 scores for most classes stabilize at high values, minor fluctuations and slower improvements occur in some classes beyond epoch 20. Together, the loss and F1 score trends indicate that the model reached its optimal generalization capability around epoch 20, making it the most reliable checkpoint for downstream tasks.

\section{Examples of Usage: Data Engineering with snail shells} \label{sec:snail_shells}
In this section, we provide the necessary information for Section~3.2 in the main draft. The snail shell dataset incorporated an 
'engineered' technique in data augmentation (see Fig.~8 in the main manuscript) that expanded an original dataset with 1.4K image/mask pairs into a dataset of size 10,000, making it a contrastingly larger dataset than the water tank section. The motivation for this dataset was to evaluate the capacity of LivePyxel's binary mask to greatly automate the gathering of data of image/mask pairs for segmentation tasks, and to evaluate its effects in the capacity of the U-Net model to accurately predict and identify among the 4 different classes of shells.

\subsection{Dataset Information}
The augmented dataset was produced via data engineering, using the masks to extract pixels from the original images as described in Section~3.2 in the main manuscript. The produced dataset was weighted and prepared as described in Section~\ref{sec:datasetPrep}. A total of 9K images were used for training, while 1K were used for validation purposes.

\begin{table}[h]
  \centering
  \begin{tabular}{|c|l|c|}
    \hline
    \textbf{Index} & \textbf{Label} & \textbf{Color (RGB)} \\ \hline
    0 & smooth\_tiger & (0, 255, 0) \\ \hline
    1 & sierpinsky    & (0, 47, 255) \\ \hline
    2 & silky         & (255, 255, 19) \\ \hline
    3 & toque         & (255, 64, 255) \\ \hline
  \end{tabular}
  \caption{Labels and corresponding RGB color codes used in the segmentation task to train the Snail-shells dataset.}
  \label{tab:label_colors_snails}
\end{table}

\subsection{Model Architecture}
\label{subsect:snailsModelArch}

The snail-shell dataset was trained on 10K images/mask pairs, each with an original resolution of 720$\times$480 pixels. Training was carried out over 10 epochs. An additional set of 100 images, not used during training or validation, were reserved for testing and detection. All training was performed using four NVIDIA V100 16 GB GPUs for a total duration of training that spanned 10 hours.

\subsection{Training}

\begin{figure}[h]
  \centering
    \includegraphics[width=\linewidth]{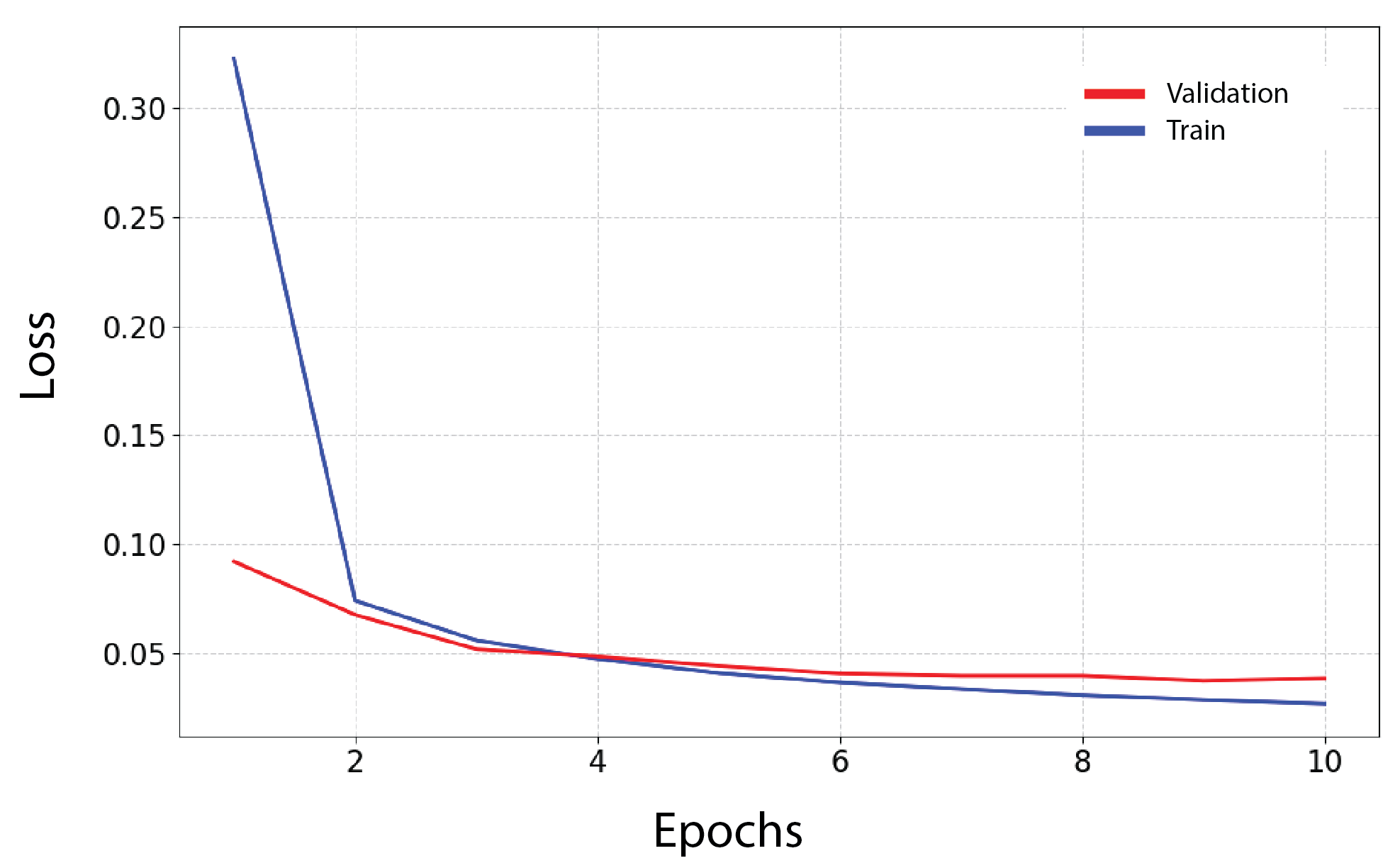}
    \caption{Cross-entropy loss curves for the training and validation sets during 10 epochs of U-Net model training on the snail-shell dataset. Both losses decrease rapidly in the initial epochs and converge steadily, indicating good generalization without signs of overfitting.}

    \label{fig:trainlossSnails}
\end{figure}

\begin{figure}[h]
  \centering
    \includegraphics[width=\linewidth]{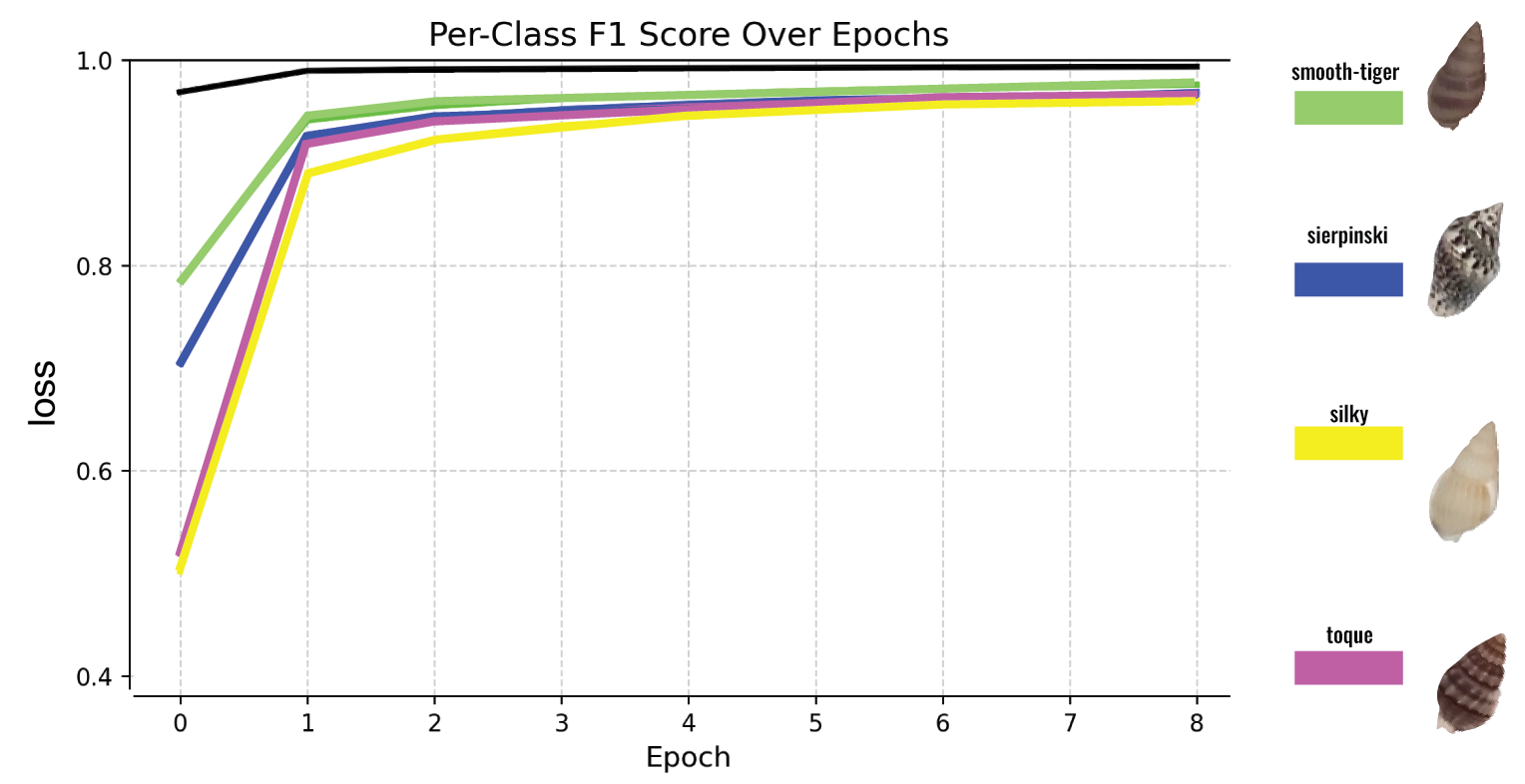}
    \caption{Per-class F1 score progression during U-Net training on the snail-shell dataset. All categories show rapid improvement in the first two epochs, stabilizing above 0.95 by epoch 3, indicating balanced and consistent segmentation performance across classes.}

    \label{fig:f1ScoresSnails}
\end{figure}

\begin{figure}[h]
  \centering
    \includegraphics[width=\linewidth]{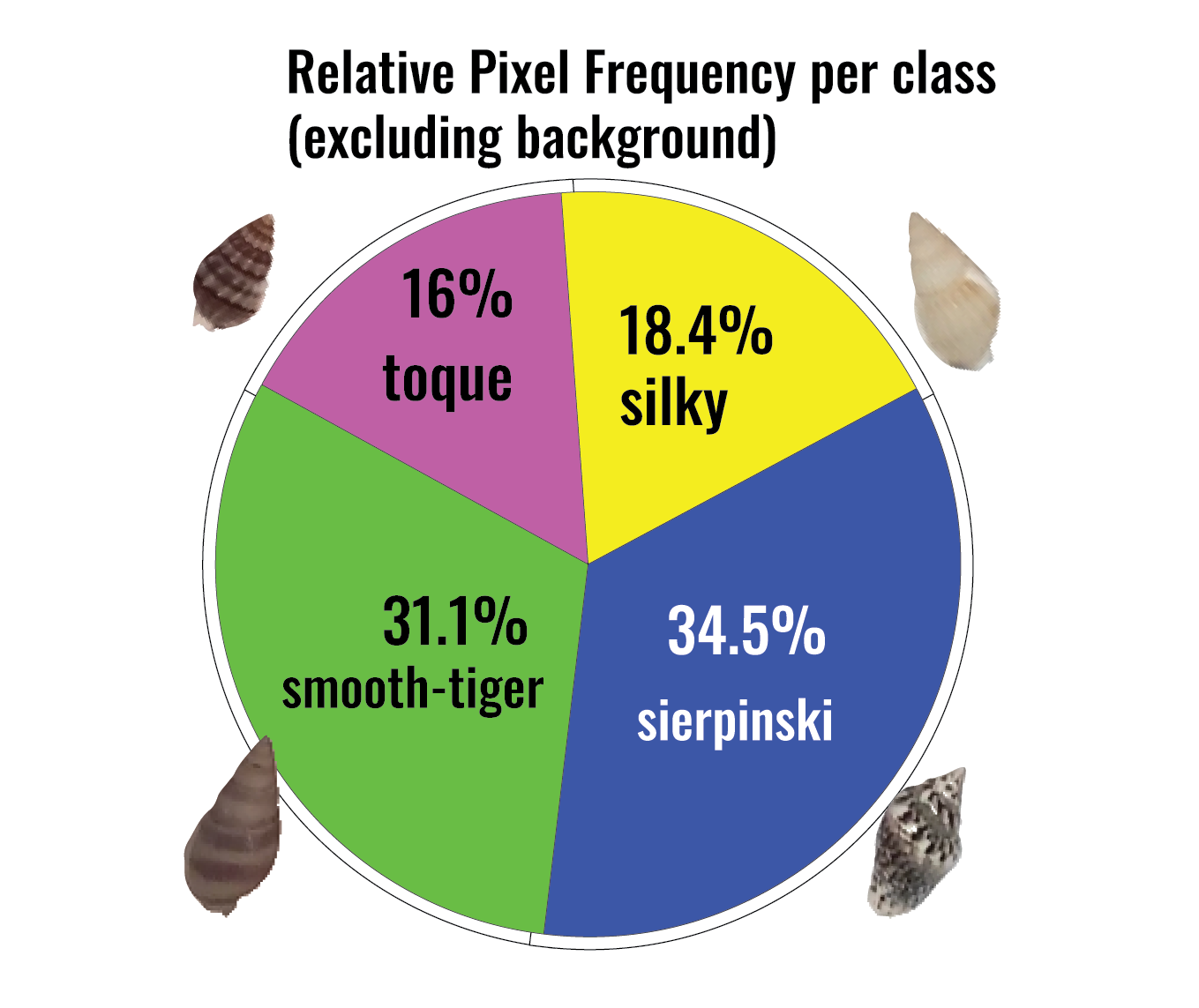}
    \caption{Relative pixel frequency distribution (excluding background) for the four classes in the snail-shell dataset. Percentages were calculated from annotation masks after normalization, highlighting the proportional representation of each shell category.}

    \label{fig:dataEng}
\end{figure}

The training and validation loss curves \ref{fig:trainlossSnails}  indicate a rapid decrease in both losses during the first few epochs, followed by a steady convergence. By epoch 10, the training loss continues to decrease slightly, while the validation loss flattens, showing minimal signs of overfitting. This suggests that the model generalizes well within this training window. Furthermore, the F1 scores per class \ref{fig:f1ScoresSnails} improve sharply in the early epochs and stabilize above 0.95 for all classes in epoch 3, demonstrating consistent and balanced performance across shell categories. Based on this performance, we selected the final model weights of epoch 8-10, where both the loss and the F1 scores suggest optimal training without overfitting.

\section{Benchmarking Annotation Tools}
\begin{figure}[h]
  \centering
    \includegraphics[width=\linewidth]{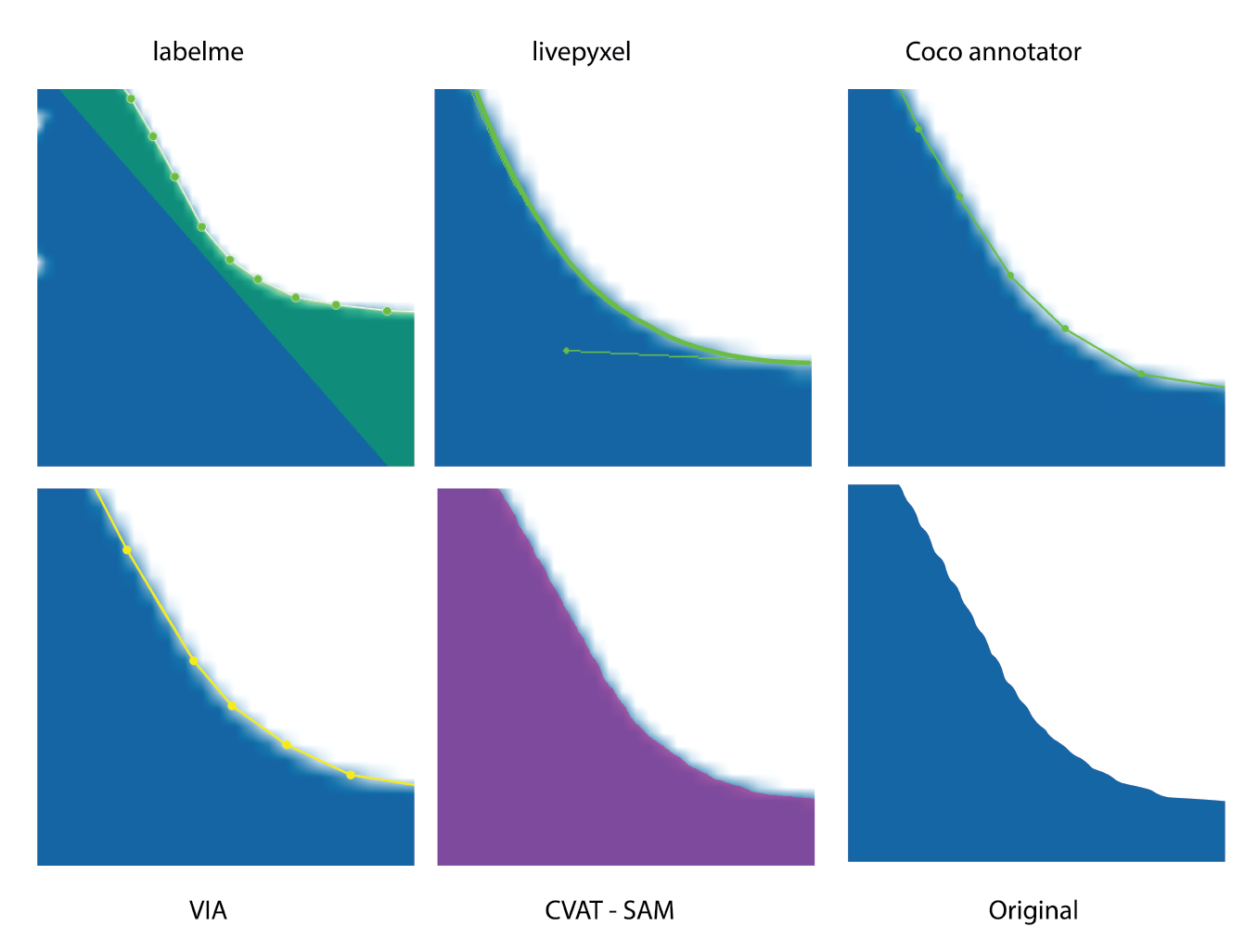}
    \caption{
    Differences in boundary smoothness and contour accuracy illustrate the variability in mask generation between traditional polygons, splines, and AI-based annotation tools; LivePyxel uses Bézier splines, CVAT uses SAM, and Labelme, COCO, and VIA use polygons.
    }
    \label{fig:ZoomToAnn}
\end{figure}

The benchmark comparing LivePyxel with other annotation tools was performed using a reference image containing five distinct polygonal shapes with varying curvature complexity (see the Original panel in Fig. 10 of the main paper).
LivePyxel achieved balanced performance, with 5.7\% false positives and 0.5\% false negatives—comparable to other software. CVAT produced the lowest overall error (2.5\% false positives, 1.3\% false negatives) by leveraging an AI segmentation tool, but required a more complex setup and data management workflow. VIA and LabelMe showed slightly higher false-positive rates, while COCO Annotator exhibited the strongest tendency toward over-segmentation.
To assess efficiency, we measured the time required to label this set of shapes across different annotation tools. The labeling times are reported in Table \ref{tab:times}, and  Fig.~\ref{fig:ZoomToAnn} displays zoomed views of the annotated boundary for five different annotation tools.
Except for LivePyxel, which used Bézier splines to better capture curved contours, polygons were used in all other software.  Notably, CVAT integrates the Segment Anything Model (SAM), which can accelerate the annotation process. 

\begin{table}[ht!]
    \centering
    \begin{tabular}{c | c}
         \textbf{Software} & \textbf{Time [s]} \\ \hline \hline
         COCO & 180 \\
         LabelMe & 254 \\
         VIA & 248\\
         CVAT$^*$ &  60\\
         LivePyxel$^\dagger$ & 173 \\
    \end{tabular}
    \caption{Time taken to annotate Fig. 10 in the main paper. $^*$CVAT incorporates SAM. $^\dagger$ Bézier splines.}
    \label{tab:times}
\end{table}

\break
\bibliography{rsc} 
\bibliographystyle{rsc} 